\documentclass[11pt]{article}
\usepackage{PRIMEarxiv}
\usepackage{etoolbox}
\makeatletter
\renewcommand{\normalsize}{%
  \@setfontsize\normalsize{11}{13.2}%
  \abovedisplayskip 8pt plus 2pt minus 4pt
  \abovedisplayshortskip 0pt plus 3pt
  \belowdisplayskip \abovedisplayskip
  \belowdisplayshortskip 5pt plus 3pt minus 3pt
}
\renewcommand{\small}{%
  \@setfontsize\small{10}{12}%
  \abovedisplayskip 7pt plus 2pt minus 4pt
  \abovedisplayshortskip 0pt plus 3pt
  \belowdisplayskip \abovedisplayskip
  \belowdisplayshortskip 4pt plus 3pt minus 3pt
}
\renewcommand{\footnotesize}{\@setfontsize\footnotesize{9}{11}}
\patchcmd{\@maketitle}
  {\vskip 0.4in \@minus 0.1in \center{ } \vskip 0.2in}
  {\vskip 18pt}
  {}{\PackageError{earn-format}{Title spacing patch failed}{}}
\makeatother
\normalsize
\usepackage[utf8]{inputenc}
\usepackage[T1]{fontenc}
\usepackage{microtype}
\usepackage[numbers,sort&compress]{natbib}
\usepackage{amsmath,amssymb}
\usepackage{amsmath,amsfonts,bm}

\def\eqref#1{equation~\ref{#1}}

\def\1{\bm{1}}

\DeclareMathAlphabet{\mathsfit}{\encodingdefault}{\sfdefault}{m}{sl}
\SetMathAlphabet{\mathsfit}{bold}{\encodingdefault}{\sfdefault}{bx}{n}

\usepackage{enumitem,wrapfig}
\usepackage{graphicx,url}
\usepackage{multirow,booktabs,array,tabularx,adjustbox}
\usepackage[table]{xcolor}
\usepackage{xspace}
\usepackage{caption}
\usepackage{algorithm,algpseudocode}
\usepackage{tcolorbox,tikz}
\usetikzlibrary{arrows.meta,positioning,calc,shapes.geometric,fit}
\usepackage[hidelinks]{hyperref}
\newcommand{\M}{\mathcal{M}}
\newcommand{\method}{EARN\xspace}
\newcolumntype{R}{>{\raggedleft\arraybackslash}X}
\title{Actions Should EARN Their Credit: Test-Time Adaptation and Route Consolidation for Frozen LLM Agents}
\author{%
{\small Bian Sun\hspace{5em}Kevin Zhai\hspace{5em}Mubarak Shah\hspace{5em}Song Wang\thanks{Corresponding author: song.wang@ucf.edu}}\\ \\
University of Central Florida, Institute of Artificial Intelligence
}
\date{}
\hypersetup{pdftitle={Actions Should EARN Their Credit: Test-Time Adaptation and Route Consolidation for Frozen LLM Agents},pdfauthor={Bian Sun; Kevin Zhai; Mubarak Shah; Song Wang}}

\begin{document}
\maketitle
\begin{abstract}
LLM agents can use feedback from repeated attempts to improve their decisions without updating model parameters. However, positive action credits can continue to favor actions that repeatedly achieve the same score, even when higher scores are achievable. Under a fixed interaction budget, trying new actions in the final attempts can also lower performance. We introduce EARN, a method that uses past experience to guide further progress and retain earlier performance through conditional reuse of effective action sequences. EARN reduces action credits for reaching scores repeatedly attained in earlier episodes and assigns larger credits to action segments that reach higher scores. These credits adjust the probabilities of LLM-proposed actions. During the final episodes, EARN directly replays stored actions when the current context matches the recorded context and the actions remain available, bypassing LLM candidate generation. We evaluate performance at the end of the interaction budget using the average score over the final four episodes. Across Jericho, ScienceWorld, and Crafter, EARN achieves the best score on this metric in 31 of 33 task--budget settings among the evaluated methods, including ties. Across the evaluated Jericho budgets, replay bypasses 78.9--86.3\% of candidate-generation calls in the final four episodes, averaged across games.
\end{abstract}

\section{Introduction}
\label{sec:introduction}

Completing interactive tasks, such as text-adventure games, multimodal open-world survival games, and virtual science experiments, typically requires an agent to perform a sequence of actions based on current observations. Because the right action is not always clear, the agent may try different actions, observe their outcomes, and adjust its choices. Specifically, an LLM agent uses the task description, current observations, and interaction records in its prompt to select the next action \citep{yao2023reactsynergizingreasoningacting}. If an attempt is unsuccessful, the agent can start again and try a different sequence. Importantly, the earlier attempt still provides useful information \citep{shinn2023reflexion}. For example, a failed attempt to open a door may reveal that the agent must first obtain a key. Such information is particularly useful in later attempts, when the agent can avoid the same mistake and focus on completing the remaining task. Thus, earlier experience can help the agent make better choices rather than repeat the same failed actions.

Existing methods help agents reuse experience without updating the LLM's parameters. Specifically, Reflexion stores written reflections to guide later attempts \citep{shinn2023reflexion}, while EvoTest revises prompts, memory, and other agent settings \citep{he2026evotest}. JitRL more directly adjusts action probabilities using returns from earlier interactions \citep{li2026justintime}. However, actions that earned points in the past do not necessarily help the agent achieve higher scores. For example, the agent may find a key and open the first door to earn points in every attempt, yet fail to complete the remaining task because it misses an important action. Although these actions remain useful, continuing to prioritize them simply because they previously earned points may lead the agent to repeat the same actions and obtain the same score. Thus, using past results effectively requires distinguishing repeated progress from progress beyond what the agent has already achieved.

Therefore, two challenges arise. \textbf{(1) Moving beyond previously achieved results.} Actions that earned points in earlier attempts may continue to receive positive credit even when they produce no further progress. For example, an agent may repeatedly open the first door but never complete the remaining task. Thus, experience reuse should distinguish previously achieved progress from new gains. \textbf{(2) Retaining earlier performance.} Choosing actions step by step based on past experience does not ensure that an earlier good result will be reproduced. Even after finding an effective action sequence, the agent may choose different actions midway through a later attempt and obtain a lower score. This matters particularly when the number of attempts is limited and only a few remain. Repeating the earlier sequence can reduce this risk, but only when its actions still apply to the current situation. For example, without the key, the agent cannot repeat an action that previously unlocked the door. Therefore, reusing an effective sequence requires checking whether the current conditions still support its execution.

\setlength{\intextsep}{8pt}
\setlength{\columnsep}{14pt}

\begin{wrapfigure}{r}{0.48\textwidth}
    \centering
\includegraphics[width=\linewidth]{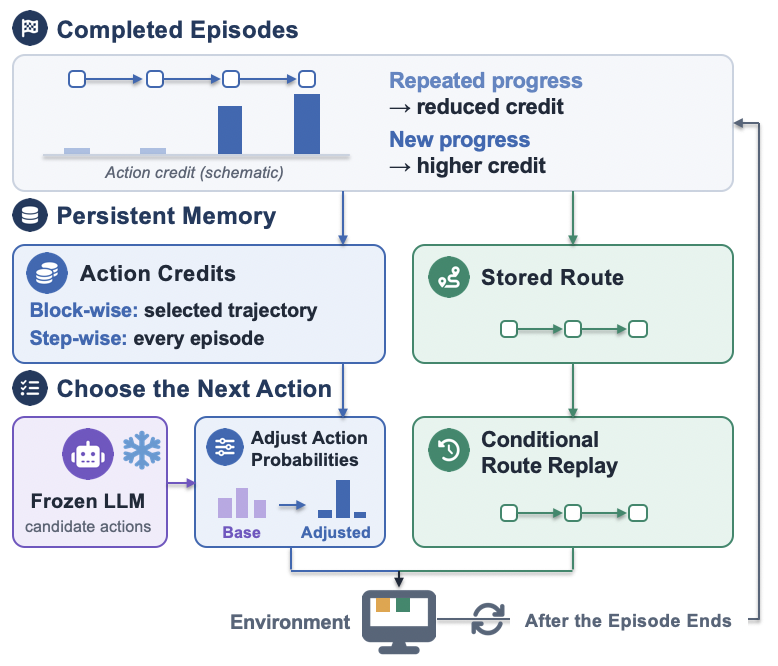}
\caption{\small \textbf{\method uses earlier scores to guide action credit assignment.}}
    \label{fig:overview}
\end{wrapfigure}

We introduce \method, a training-free method that uses persistent memory to adjust action probabilities and conditionally reuse a previously effective action sequence (Figure~\ref{fig:overview}). We group consecutive episodes into blocks and select the highest-scoring trajectory from each block. From this trajectory, \method assigns numerical credits to actions leading up to a score increase, using these credits to guide later choices. Actions receive less credit when the resulting score does not exceed a level reached more than once in earlier episodes. To also use information from the other trajectories, \method computes step-wise credits after every episode. During action selection, it combines the two credit estimates and reduces the block-wise estimate when few records match the current situation and candidate action, or when their credit values vary substantially. Separately, \method stores a route containing an ordered action sequence and the situation recorded before each action. During the final episodes, the agent directly executes stored actions while the recorded and current situations match and the actions remain available. Otherwise, it resumes memory-guided action selection.

We evaluate whether this combination of memory-guided decisions and conditional route reuse improves performance in the final episodes across Jericho \citep{hausknecht2020interactive}, ScienceWorld \citep{wang2022scienceworld}, and Crafter \citep{hafner2022benchmarking}. Our evaluation covers three aspects: final-four average scores relative to existing baselines, retention of previously achieved scores and inference savings when routes remain executable, and adaptation when task instances change. Together, these experiments assess whether EARN can improve later decisions and reuse accumulated experience under both repeated and changing execution conditions. Our contributions are threefold:

\begin{itemize}[leftmargin=0.35cm, itemsep=0pt, topsep=1.5pt]
\item We introduce \method, a training-free method that combines progress-relative block-wise credit attribution from selected trajectories with step-wise feedback from every episode to adjust candidate-action probabilities through support-controlled fusion without parameter updates.

\item We develop conditional route replay to complement credit-guided action selection. It preserves an applicable route's action order and reduces LLM candidate-generation calls by directly executing stored actions. When replay stops, memory-guided action selection resumes.

\item We evaluate \method across three interactive benchmarks. Among the compared methods, \method achieves the highest or tied-highest final-four average episode score in 14 of 15 Jericho game--budget settings, 14 of 15 ScienceWorld task--budget settings, and all three Crafter budgets.
\end{itemize}

\section{Related Work}

\textbf{Test-time adaptation of LLM agents.} Existing methods differ in what they retain from earlier episodes and how the agent uses that information in later decisions. Some methods store text derived from earlier episodes: Reflexion~\citep{shinn2023reflexion} stores verbal reflections on task feedback, CLIN~\citep{majumder2024clin} records causal relations between actions and observed effects, and ReasoningBank~\citep{ouyang2026reasoningbank} distills reasoning strategies from successful and failed trajectories. Beyond updating memory, EvoTest~\citep{he2026evotest} revises the agent's prompt, hyperparameters, and tool-use routines. Other methods store numerical estimates for actions. REMEMBERER~\citep{zhang2023rememberer} retrieves past actions and their estimated scores to guide action selection through the prompt. JitRL~\citep{li2026justintime} is the closest prior method to \method. It uses an LLM evaluator to assign numerical feedback to actions, computes discounted returns, and averages retrieved returns for each candidate action to adjust action probabilities. \method instead uses a score repeatedly attained in earlier episodes as a reference for assigning action credits and discounts credits for score increases up to that reference. It combines block-wise credits from one selected trajectory per block with step-wise credits from every episode to adjust action probabilities.

\textbf{Reusing behavior from earlier episodes.} Prior systems represent behavior from earlier interactions as trajectories, textual guidance, workflows, or executable skills. ExpeL~\citep{zhao2024expel}, Synapse~\citep{zheng2024synapse}, and Self-Generated In-Context Examples~\citep{sarukkai2025selfgenerated} use earlier trajectories as in-context examples from which the LLM generates new actions. AutoGuide~\citep{fu2024autoguide} derives textual guidelines from trajectories and pairs each guideline with a description of the context in which it applies. Agent Workflow Memory~\citep{wang2025awm} derives reusable workflows from trajectories and includes the workflows in prompts for later tasks. Voyager~\citep{wang2024voyager} stores executable code skills that can be retrieved and composed. \method stores a route containing recorded actions and their preceding execution contexts, then directly executes those actions in order while execution checks pass.
\section{Problem Formulation}
\label{sec:problem}

\textbf{Preliminaries.} We consider interactive tasks in which an agent takes a sequence of actions to achieve an objective, and the environment provides observations and scores that indicate task progress. Because a single attempt may leave parts of the task unfinished, attempts allow the agent to try alternative actions and use feedback from earlier attempts. We therefore consider a session of $K$ episodes of the same task, with a fixed objective and scoring criterion. Each episode $k$ begins with an environment reset to initial configuration $v_k$, which may vary across episodes, and ends upon environment termination or after $H$ steps, giving a length $T_k\le H$. At step $t$, the agent receives execution context $s_{k,t}$, containing the current observation and supplementary information such as available actions, and executes action $a_{k,t}$. The environment then returns feedback $y_{k,t}$, including the next context, the score change, and a termination indicator. Let $S_{k,t}$ denote the running score after action $a_{k,t}$ and $S_{k,0}$ the initial score. The score change at step $t$ is $S_{k,t}-S_{k,t-1}$.

\textbf{Memory Update and Action Selection.} The LLM parameters $\theta$ remain fixed throughout the session. Persistent memory $\M_k$ retains information from episodes $1,\ldots,k-1$, with $\M_1=\varnothing$. The agent constructs prompt $x_{k,t}$ from the task specification, current context, within-episode history, and retrieved memory. An action-selection rule $\Pi_\theta$ uses both the prompt and persistent memory:
\begin{equation}
a_{k,t}\sim\Pi_\theta(\cdot\mid x_{k,t},\M_k).
\label{eq:memory_conditioned_action}
\end{equation}
The trajectory $\tau_k$ records the ordered sequence of contexts, actions, and feedback from episode $k$:
\begin{equation}
\tau_k=\bigl(s_{k,t},a_{k,t},y_{k,t}\bigr)_{t=1}^{T_k}.
\label{eq:episode_trajectory}
\end{equation}
The task's scoring criterion assigns episode score $S_k$ to the completed trajectory. An action credit is instead a numerical value constructed from interaction feedback to guide action selection in later episodes. After each episode, a memory-update rule $U$ produces the memory for the next episode:
\begin{equation}
\M_{k+1}=U(\M_k,\tau_k,S_k).
\label{eq:memory_update}
\end{equation}

\textbf{Goal.} We seek a memory-update rule $U$ and an action-selection rule $\Pi_\theta$ that maximize the expected average score over the final $W$ episodes, where $1\le W\le K$, subject to the episode budget $K$ and per-episode step limit $H$:
\begin{equation}
J_{\mathrm{term}}
=
\mathbb E_{\mathrm{agent},\,\mathrm{environment}}
\left[
\frac{1}{W}\sum_{k=K-W+1}^{K}S_k
\right].
\label{eq:terminal_objective}
\end{equation}

\section{Method}
\begin{figure}[!hbp]
    \centering
    \includegraphics[width=\linewidth]{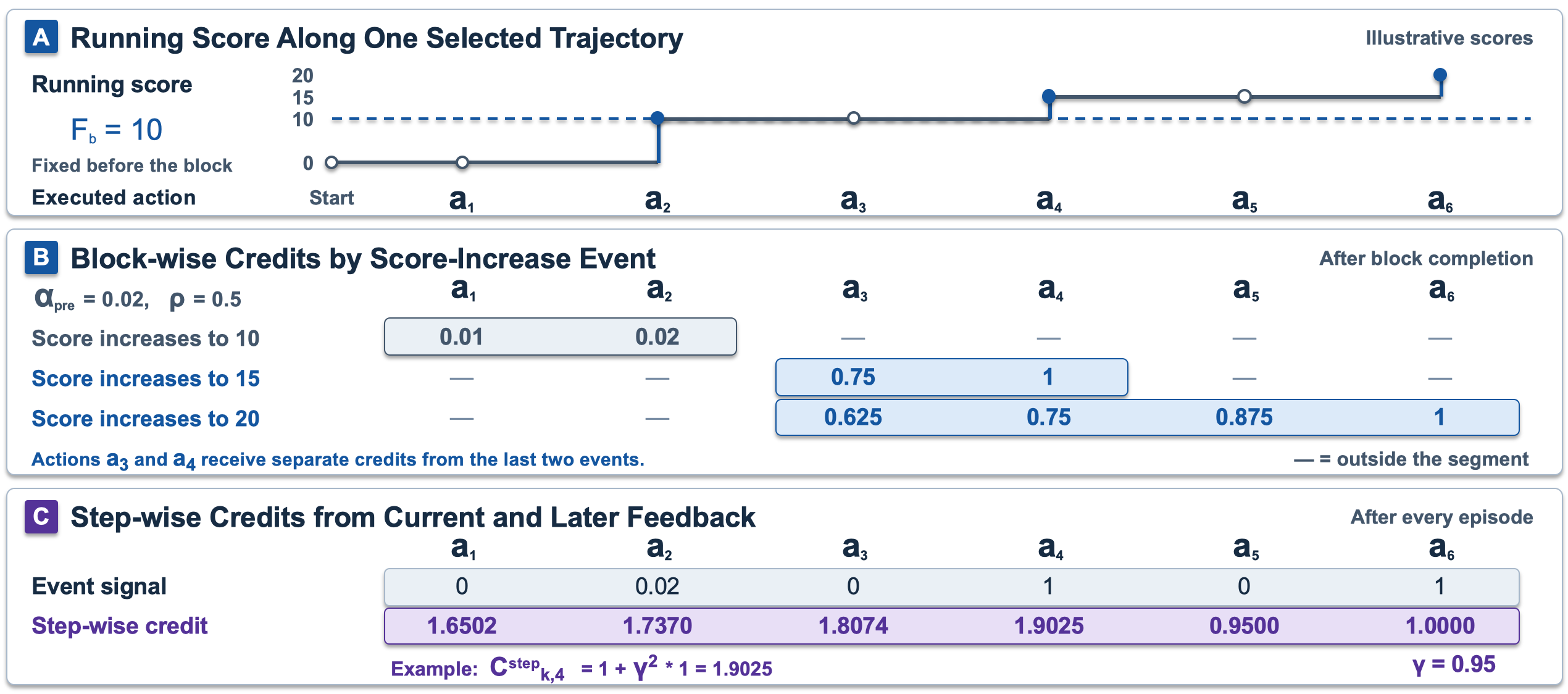}
\caption{\footnotesize \textbf{An illustrative example of EARN's} block-wise and step-wise credit attribution. Environment scores are illustrative. Both attribution rules use a frontier of 10 in this example; the step-wise calculation includes only score-increase signals.}
    \label{fig:method-overview}

\end{figure}

\method uses scores repeatedly attained in earlier episodes as a reference for assigning action credits, discounting credits for reaching those scores again. The memory-update rule $U$ adds step-wise credits after every episode and block-wise credits from score-increase events after selecting one trajectory from each block. The action-selection rule $\Pi_\theta$ retrieves credits to adjust candidate-action probabilities and conditionally replays a stored route. Figure~\ref{fig:method-overview} illustrates both credit assignments on one trajectory. Algorithm~\ref{alg:earn}(Appendix~\ref{app:algo}) summarizes our methods.

\subsection{Learning: Action Credit and Persistent Memory}
\label{sec:learning}

\textbf{Block-wise Credit Attribution.} To prioritize experience with greater task progress, we compare trajectories within non-overlapping blocks of $B$ consecutive acquisition episodes and select one per block, primarily by the highest running score reached. This provides several outcomes to compare while allowing later experience to contribute without surpassing the best score of the entire session. To give less credit for progress already achieved, we maintain a progress frontier $F_b$ using route scores observed multiple times in recent episodes. The frontier starts at zero, never decreases, and is fixed at the start of each block. A score-increase event $e$ occurs when $S_{k,t_e}>S_{k,t_e-1}$; let $S_e=S_{k,t_e}$. Because earlier actions may contribute to this increase, we credit a segment rather than only the scoring action. For $S_e\le F_b$, we use a local segment ending at $t_e$. Otherwise, we use the segment after first reaching or exceeding $F_b$ through $t_e$, covering the actions leading to progress beyond the frontier. Appendix~\ref{app:block_frontier} gives the selection, frontier-update, and segment details.

For a segment of $L_e$ actions, let $c_{e,i}$ denote the credit assigned to the action at position $i$. We assign larger credits to actions closer to the score-increase event and discount credits for repeated progress. The interpolation endpoints $\ell_e$ and $h_e$ depend on the discount coefficient $\alpha_{\mathrm{pre}}$ for repeated progress and the credit offset $\rho$ for progress beyond $F_b$:
\begin{equation}
c_{e,i}
=\left(1-\frac{i}{L_e}\right)\ell_e+\frac{i}{L_e}h_e,
\qquad
(\ell_e,h_e)=
\begin{cases}
(0,\alpha_{\mathrm{pre}}), & S_e\le F_b,\\
(\rho,1), & S_e>F_b,
\end{cases}
\quad i=1,\ldots,L_e.
\label{eq:selected_credit}
\end{equation}
Credit increases linearly toward the scoring action. For repeated progress, it receives $\alpha_{\mathrm{pre}}$; for new progress, it receives 1. The offset $\rho$ gives preceding actions positive credit in new-progress segments.

\textbf{Step-wise Credit Attribution.} To provide timely feedback between block-wise updates and retain information from unselected trajectories, we compute step-wise credits after every episode, including replay episodes. For each step $t$, we construct a numerical signal $\tilde r_{k,t}$ from environment feedback and interaction history. A score-increase event contributes $1$ if the running score after the event exceeds the progress frontier, and a discounted positive signal otherwise. Here, we use the frontier before updating memory from the current episode. We additionally give small bonuses for previously unobserved outcomes and apply penalties for detected terminal failures. Using temporal discount factor $\gamma$, we accumulate these signals backward through the episode to compute step-wise credit $C^{\mathrm{step}}_{k,t}$ for action $a_{k,t}$:
\begin{equation}
    C^{\mathrm{step}}_{k,t} =\sum_{u=t}^{T_k}\gamma^{u-t}\tilde r_{k,u}.
\label{eq:step_credit}
\end{equation}
Here, $\gamma$ controls the decay of event feedback as it is propagated backward to preceding actions. An action can receive credit from subsequent events even if it produces no immediate score increase. The resulting credits are stored after each episode, without additional LLM evaluation calls.

\textbf{Persistent Memory.} Persistent memory $\M_k$ stores a route $\mathcal R_k^\star$ and action-credit records $\mathcal E_k$. A route is an ordered action sequence extracted from an episode trajectory, together with the execution context recorded before each action. Each credit record contains an execution context, an action, an assigned credit, and the credit source. Step-wise records are added after every episode; block-wise records from score-increase events are added after selecting a trajectory at block completion. Route storage is independent of block-wise selection: after each episode, the agent considers the new trajectory for route retention based on performance and executability. Details are in Appendix~\ref{sec:method_details}.

\subsection{Adaptation: Adjusting Candidate-Action Probabilities}
\label{sec:evolution}

Given execution context $s=s_{k,t}$, the agent incorporates retrieved memory into prompt $x_{k,t}$. The frozen LLM then proposes candidate actions and reports confidence scores, which we normalize into a base distribution $p^0(a\mid s)$ over the candidate set $\mathcal A(s)$. Additional memory candidates, when enabled, are incorporated as described in Appendix~\ref{app:action_retrieval}.

\textbf{Credit Estimates.} \method retrieves block-wise and step-wise credit records from persistent memory $\M_k$. Semantic retrieval can supplement insufficient exact matches; Appendix~\ref{app:action_retrieval} describes this procedure. For block-wise records matching context $s$ and candidate action $a$ exactly, let $n_{\mathrm{blk}}$, $\mu_{\mathrm{blk}}$, and $\sigma_{\mathrm{blk}}^2$ denote their count, mean, and sample variance. These statistics depend on $(s,a)$; we omit the arguments for brevity. To moderate estimates based on limited or inconsistent credit records, we shrink the block-wise estimate toward zero using count-based scaling and a variability correction:
\begin{equation}
r_{\mathrm{blk}}(s,a)=\frac{n_{\mathrm{blk}}}{n_{\mathrm{blk}}+\nu},
\qquad
Q_{\mathrm{blk}}(s,a)
=r_{\mathrm{blk}}(s,a)\operatorname{sgn}(\mu_{\mathrm{blk}})
\left[
|\mu_{\mathrm{blk}}|
-\xi\sqrt{\frac{\sigma_{\mathrm{blk}}^2}{n_{\mathrm{blk}}}}
\right]_+.
\label{eq:block_value}
\end{equation}
Here, $[x]_+=\max(0,x)$, $\nu$ controls count-based scaling, and $\xi$ controls the variability correction. Without matching records, $r_{\mathrm{blk}}=Q_{\mathrm{blk}}=0$. Each retrieved step-wise credit is scaled according to its episode's replay status and whether the running score before or after the action exceeded that episode's initial progress frontier. The estimate $Q_{\mathrm{step}}(s,a)$ is the mean scaled credit over records matching action $a$. We center each estimate against its channel's baseline: $A_{\mathrm{blk}}=Q_{\mathrm{blk}}-V_{\mathrm{blk}}$ and $A_{\mathrm{step}}=Q_{\mathrm{step}}-V_{\mathrm{step}}$. Here, $V_{\mathrm{blk}}(s)$ is the mean block-wise estimate over candidate actions weighted by $r_{\mathrm{blk}}$, and $V_{\mathrm{step}}(s)$ is the mean scaled credit over all retrieved step-wise records. These centered estimates express each candidate's value relative to its channel's baseline.

\textbf{Combining Credit Estimates.} Because block-wise selection retains only one trajectory per block, step-wise credits use feedback from every episode, including those not selected. When matching block-wise records are sparse, we retain a larger coefficient on the step-wise estimate. As block-wise support grows, we reduce this coefficient to place greater emphasis on credits from selected trajectories. We therefore combine the centered estimates as:
\begin{equation}
A_{\mathcal M}(s,a)
=g_{\mathrm{blk}}A_{\mathrm{blk}}(s,a)
+\lambda_{\mathrm{step}}[1-\eta g_{\mathrm{blk}}]_+
\,g_{\mathrm{step}}q_{\mathrm{step}}A_{\mathrm{step}}(s,a).
\label{eq:fusion}
\end{equation}
The gates $g_{\mathrm{blk}},g_{\mathrm{step}}\in[0,1]$ regulate each channel's contribution according to retrieved support. The coefficient $\lambda_{\mathrm{step}}$ sets the step-wise scale, $\eta$ controls its suppression as $g_{\mathrm{blk}}$ increases, and $q_{\mathrm{step}}$ is the largest record-scaling factor among matching step-wise records. The gates and $q_{\mathrm{step}}$ depend on $(s,a)$; their arguments are omitted. Step-wise feedback supplements sparse block-wise evidence. As block-wise support grows, the suppression factor reduces the step-wise contribution to the combined estimate before standardization. Appendix~\ref{app:step_value} specifies the record-scaling factors.

\textbf{Probability Adjustment.} We standardize $A_{\mathcal M}(s,a)$ over the candidate set using its mean and population standard deviation, then clip the result to $[-\zeta,\zeta]$, obtaining $\widehat A_{\mathcal M}(s,a)$. With adjustment strength $\beta>0$, the adjusted distribution is
\begin{equation}
p_{\mathcal M}(a\mid s)
=
\frac{
p^0(a\mid s)\exp\!\left(\beta\widehat A_{\mathcal M}(s,a)\right)
}{
\sum_{a'\in\mathcal A(s)}
p^0(a'\mid s)\exp\!\left(\beta\widehat A_{\mathcal M}(s,a')\right)
}.
\label{eq:probability_adjustment}
\end{equation}
This adjustment shifts the base probability ratio in favor of candidates with higher standardized credit estimates, while retaining the LLM's base probabilities as a factor. The parameter $\zeta>0$ bounds the standardized estimates. Appendix~\ref{app:probability_adjustment} describes the numerical handling.
\subsection{Conditional Route Replay}

Block-wise and step-wise credits adjust candidate-action probabilities, but do not ensure that the agent follows the action order in a stored route $\mathcal R_k^\star$. The agent therefore maintains a replayable route separately and conditionally follows its action order when execution checks pass.

Let $K_c$ denote the number of final episodes scheduled for route replay, with $z_k=\mathbb I\{k>K-K_c\}$ indicating replay eligibility. The evaluation window $W$ independently specifies which episode scores enter $J_{\mathrm{term}}$. Let $m$ index the next action $a_m^\star$ in $\mathcal R_k^\star$. The indicator $\chi_{k,t}\in\{0,1\}$ equals one when replay remains active in the episode, a route step remains, the current execution context matches its recorded context, and the action passes availability and terminal-failure checks. The operator $\operatorname{Select}$ follows the configured action-selection rule. The agent executes
\begin{equation}
a_{k,t}=
\begin{cases}
a_m^\star, & z_k\chi_{k,t}=1,\\[3pt]
\operatorname{Select}\!\left(
p_{\mathcal M}(\cdot\mid s_{k,t})\right),
& \text{otherwise}.
\end{cases}
\label{eq:conditional_replay}
\end{equation}
Each replayed action advances the route pointer and bypasses LLM candidate generation, while still counting as an environment interaction. A failed execution check or an exhausted route ends replay for the current episode. The agent then resumes candidate generation and uses the block-wise and step-wise credit estimates to adjust action probabilities.

\section{Experiments}
We organize our experiments around three questions on memory-guided test-time adaptation and final-stage consolidation in LLM agents: \textbf{Q1: Performance in the final episodes.} Does EARN improve performance in the final episodes over the compared baselines across interactive environments? \textbf{Q2: Performance Retention and Inference Savings.} When stored routes are broadly applicable, how reliably does EARN reproduce its best acquired performance, and how many LLM candidate-generation calls does conditional consolidation save?  \textbf{Q3: Experience Transfer.} When stored routes are inapplicable, how does EARN use accumulated experience to adapt its decisions and maintain performance in the final episodes?
\subsection{Experimental Setup}
\textbf{Environments and Tasks. }Experiments span three benchmarks with complementary demand profiles. \textbf{Jericho} ~\citep{hausknecht2020interactive} offers parser-based interactive fiction; we adopt the Focus-5 subset (DeepHome, Detective, Zork~I, Zork~III, ZTUU) and run lifelong sessions of 30$\times$30, 50$\times$50, and 70$\times$70 (episodes $\times$ steps per episode). \textbf{ScienceWorld}~\citep{wang2022scienceworld} is evaluated on five tasks, with at most 100 steps per episode and session lengths of 10, 20, and 30 episodes. Task variations alter objects, materials, or environmental configurations across episodes, requiring stored experience to be checked for applicability to the current instance. \textbf{Crafter}~\citep{hafner2022benchmarking} is an open-ended survival environment evaluated in a multimodal setting using visual observations and textual context. Performance is measured by the number of distinct achievements unlocked (0--22), with at most 100 steps per episode and session lengths of 10, 20, and 30 episodes.

\textbf{Compared Approaches.} All approaches share an identical frozen W4A16-quantized backbone Gemma-4-31B-Instruct \citep{gemmateam2026gemma4technicalreport}, served locally via vLLM \citep{kwon2023efficientmemorymanagementlarge}. Baselines retain their native decoding, memory, retrieval settings. \method uses an action-generation temperature of 0.8 across all three benchmarks. \method uses all-MiniLM-L6-v2 \citep{wang2020minilmdeepselfattentiondistillation} embeddings for semantic retrieval in Jericho and ScienceWorld, exact-context retrieval for action evidence in Crafter. We compare against five baselines that reuse experience while keeping the LLM weights fixed: (1)~\textbf{Reflexion}~\citep{shinn2023reflexion}, which appends verbal self-critiques of past failures to the prompt; (2)~\textbf{AWM}~\citep{wang2025awm}, which distills successful episodes into persisted reusable workflows; (3)~\textbf{FIFO Memory}, which prepends full episode transcripts under a first-in-first-out context budget; (4)~\textbf{EvoTest}~\citep{he2026evotest}, which evolves prompts, memory entries, and decoding hyperparameters across episodes; and (5)~\textbf{JitRL}~\citep{li2026justintime}, which converts returns of past trajectories into advantages that rescale candidate-action probabilities.

\textbf{Protocol and Metrics.} All methods are evaluated under the same episode and step budgets. We report \textbf{Last-4}, $S_{\mathrm{Last\text{-}4}}=\frac{1}{4}\sum_{k=K-3}^{K}S_k$, and \textbf{Last}, $S_{\mathrm{Last}}=S_K$. For Jericho, we additionally report \textbf{Retention}, $R_{\mathrm{ret}}=\frac{1}{4}\sum_{k=K-3}^{K} \mathbf{1}[S_k=S^\star \land S^\star>0]$, where $S^\star=\max_{1\le j\le K}S_j$ is the run's peak score. Retention measures the fraction of final-window episodes matching a positive peak and is zero for runs without one. EARN enables conditional route replay during the final-four consolidation window. More implementation details are provided in Appendices~\ref{app:final_window}.

\subsection{Main Results}

\textbf{Performance in the final episodes (Q1).} EARN achieves the highest or tied Last-4 score in 14 of 15 Jericho game--budget settings (Table~\ref{tab:jericho-last4}) and the best score in 14 of 15 ScienceWorld settings (Table~\ref{tab:scienceworld-results}), including ties. Crafter results evaluate its performance in multimodal interaction (Table~\ref{tab:crafter-results}). Across these settings, the results support the value of selectively reusing accumulated experience for terminal performance, both when stored routes remain applicable and when decisions require adaptation to the current context. Larger interaction budgets do not uniformly yield higher scores, highlighting that how experience is used matters alongside how much is collected. Interaction creates opportunities for progress, but does not by itself ensure that experience will be retained and effectively reused. The comparison highlights the importance of experience selection and context-aware reuse, alongside exploration, for achieving strong performance near the budget endpoint.

\begin{table}[!hbp]
\centering
\footnotesize
\caption{Jericho results under different episode/step budgets. Last-4 denotes the average score over the final four episodes, and Last denotes the final episode's score ($\uparrow$). Bold values indicate the best results within each budget.}
\vspace{1mm}
\label{tab:jericho-last4}

\setlength{\tabcolsep}{3pt}
\renewcommand{\arraystretch}{1.08}
\setlength{\aboverulesep}{2pt}
\setlength{\belowrulesep}{2pt}
\begin{tabularx}{\linewidth}{@{}cl*{10}{R}@{}}
\toprule
\multirow{2}{*}{\shortstack{\textbf{Episodes}\\\textbf{/ Steps}}}
& \multirow{2}{*}{\textbf{Method}}
& \multicolumn{2}{c}{\textbf{DeepHome}}
& \multicolumn{2}{c}{\textbf{Detective}}
& \multicolumn{2}{c}{\textbf{Zork I}}
& \multicolumn{2}{c}{\textbf{Zork III}}
& \multicolumn{2}{c}{\textbf{ZTUU}} \\
\cmidrule(lr){3-4}\cmidrule(lr){5-6}\cmidrule(lr){7-8}
\cmidrule(lr){9-10}\cmidrule(lr){11-12}
& & \textbf{Last-4} & \textbf{Last}
& \textbf{Last-4} & \textbf{Last}
& \textbf{Last-4} & \textbf{Last}
& \textbf{Last-4} & \textbf{Last}
& \textbf{Last-4} & \textbf{Last} \\
\midrule
\multirow{6}{*}{30/30}
& Memory    & 13.8 & 14.0 & 70.0 & 80.0 & 5.0 & 5.0 & 1.8 & 2.0 & 3.8 & 5.0 \\
& Reflexion & 6.8 & 7.0 & 20.0 & 20.0 & 2.5 & 0.0 & 0.8 & 1.0 & 0.0 & 0.0 \\
& AWM       & 14.0 & 14.0 & 50.0 & 50.0 & 10.0 & 10.0 & 0.0 & 0.0 & 0.0 & 0.0 \\
& EvoTest   & 14.8 & 15.0 & 82.5 & 90.0 & 40.0 & 40.0 & 2.3 & 2.0 & 2.5 & 5.0 \\
& JitRL     & 19.0 & 18.0 & 172.5 & \textbf{190.0} & 42.3 & \textbf{50.0} & 2.5 & 2.0 & 6.3 & 10.0 \\
\rowcolor{blue!6}
& \textbf{Ours} & \textbf{19.3} & \textbf{20.0} & \textbf{190.0} & \textbf{190.0} & \textbf{47.5} & \textbf{50.0} & \textbf{4.0} & \textbf{4.0} & \textbf{20.8} & \textbf{22.0} \\
\midrule
\multirow{6}{*}{50/50}
& Memory    & 13.5 & 14.0 & 90.0 & 110.0 & 41.3 & 45.0 & 3.0 & 3.0 & 7.5 & 10.0 \\
& Reflexion & 5.0 & 1.0 & 17.5 & 20.0 & 2.5 & 0.0 & 0.5 & 0.0 & 0.0 & 0.0 \\
& AWM       & 15.0 & 15.0 & 72.5 & 80.0 & 10.0 & 10.0 & 1.0 & 1.0 & 0.0 & 0.0 \\
& EvoTest   & 15.0 & 12.0 & 310.0 & 310.0 & 41.0 & 40.0 & 2.5 & 3.0 & 8.8 & 10.0 \\
& JitRL     & 14.8 & 19.0 & 310.0 & 310.0 & 47.5 & 50.0 & 3.0 & 3.0 & 0.0 & 0.0 \\
\rowcolor{blue!6}
& \textbf{Ours} & \textbf{23.0} & \textbf{23.0} & \textbf{340.0} & \textbf{340.0} & \textbf{54.0} & \textbf{54.0} & \textbf{5.0} & \textbf{5.0} & \textbf{27.0} & \textbf{27.0} \\
\midrule
\multirow{6}{*}{70/70}
& Memory    & 16.3 & 13.0 & 252.5 & 300.0 & 5.0 & 5.0 & 2.8 & 3.0 & \textbf{20.0} & 20.0 \\
& Reflexion & 4.8 & 1.0 & 17.5 & 20.0 & 5.0 & 0.0 & 0.5 & 1.0 & 0.0 & 0.0 \\
& AWM       & 27.0 & 27.0 & 50.0 & 50.0 & 0.0 & 0.0 & 3.0 & 3.0 & 5.0 & 5.0 \\
& EvoTest   & 51.5 & \textbf{64.0} & 315.0 & 310.0 & 42.0 & 40.0 & 2.8 & 3.0 & 10.0 & 10.0 \\
& JitRL     & 20.0 & 20.0 & 300.0 & 300.0 & 41.0 & 55.0 & 2.8 & 4.0 & 0.0 & 0.0 \\
\rowcolor{blue!6}
& \textbf{Ours} & \textbf{64.0} & \textbf{64.0} & \textbf{320.0} & \textbf{320.0} & \textbf{59.0} & \textbf{59.0} & \textbf{5.0} & \textbf{5.0} & 19.5 & \textbf{27.0} \\
\bottomrule
\end{tabularx}

\end{table}

\begin{table}[!t]
\centering
\footnotesize
\caption{ScienceWorld results across task variations, i.e., instances of the same task with different objects, materials, or environmental configurations. Last-4 denotes the mean solving score over the final four episodes, and Last denotes the final episode's score ($\uparrow$). Each episode allows 100 steps.}
\vspace{1mm}

\label{tab:scienceworld-results}
\setlength{\tabcolsep}{4pt}
\renewcommand{\arraystretch}{1.08}
\setlength{\aboverulesep}{2pt}
\setlength{\belowrulesep}{2pt}

\newcommand{\taskhead}[1]{%
  \multicolumn{2}
  {>{\centering\arraybackslash}p{\dimexpr0.13\textwidth+2\tabcolsep\relax}}
  {\bfseries #1}}

\resizebox{\textwidth}{!}{%
\begin{tabular}{
  @{}cl*{10}{>{\raggedleft\arraybackslash}p{0.065\textwidth}}@{}
}
\toprule
\multirow{2}{*}{\textbf{Episodes}}
& \multirow{2}{*}{\textbf{Method}}
& \taskhead{Conductivity}
& \taskhead{Unknown Conductivity}
& \taskhead{Plant}
& \taskhead{Power}
& \taskhead{Melting Known} \\
\cmidrule(lr){3-4}\cmidrule(lr){5-6}
\cmidrule(lr){7-8}\cmidrule(lr){9-10}
\cmidrule(lr){11-12}
& & \textbf{Last-4} & \textbf{Last}
  & \textbf{Last-4} & \textbf{Last}
  & \textbf{Last-4} & \textbf{Last}
  & \textbf{Last-4} & \textbf{Last}
  & \textbf{Last-4} & \textbf{Last} \\
\midrule
\multirow{6}{*}{10}
& Memory    & 19.0 & 19.0 & 13.3 & 10.0 & 11.3 & 13.0 & \textbf{58.0} & \textbf{58.0} & \textbf{100.0} & \textbf{100.0} \\
& AWM       & 19.0 & 19.0 & 14.5 & 19.0 & 11.0 & 13.0 & 20.0 & 20.0 & 93.3 & \textbf{100.0} \\
& Reflexion & 10.0 & 10.0 & 10.0 & 10.0 & 3.3 & 1.0 & 10.0 & 10.0 & 86.5 & 73.0 \\
& EvoTest   & 27.8 & 29.0 & 24.3 & 29.0 & 11.8 & 13.0 & \textbf{58.0} & \textbf{58.0} & 93.3 & \textbf{100.0} \\
& JitRL     & 47.5 & 60.0 & 60.0 & 60.0 & \textbf{34.8} & \textbf{37.0} & 22.8 & 0.0 & 84.8 & \textbf{100.0} \\
& \textbf{Ours}
& \textbf{87.3} & \textbf{83.0} & \textbf{82.0} & \textbf{83.0}
& \textbf{34.8} & \textbf{37.0} & \textbf{58.0} & \textbf{58.0}
& \textbf{100.0} & \textbf{100.0} \\
\midrule
\multirow{6}{*}{20}
& Memory    & 17.8 & 33.0 & 16.5 & 14.0 & 10.8 & 11.0 & \textbf{58.0} & \textbf{58.0} & \textbf{100.0} & \textbf{100.0} \\
& AWM       & 19.0 & 19.0 & 12.3 & 29.0 & 10.3 & 11.0 & 25.0 & 25.0 & 71.3 & \textbf{100.0} \\
& Reflexion & 8.8 & 10.0 & 8.8 & 10.0 & 3.5 & 5.0 & 10.0 & 10.0 & 93.3 & 73.0 \\
& EvoTest   & 31.0 & 29.0 & 22.8 & 29.0 & 13.0 & 13.0 & \textbf{58.0} & \textbf{58.0} & \textbf{100.0} & \textbf{100.0} \\
& JitRL     & 60.0 & 60.0 & 60.0 & 60.0 & \textbf{33.8} & \textbf{33.0} & 24.0 & 5.0 & 76.5 & \textbf{100.0} \\
& \textbf{Ours}
& \textbf{74.0} & \textbf{69.0} & \textbf{77.5} & \textbf{79.0}
& \textbf{33.8} & \textbf{33.0} & 55.5 & 53.0
& \textbf{100.0} & \textbf{100.0} \\
\midrule
\multirow{6}{*}{30}
& Memory    & 18.0 & 10.0 & 18.0 & 24.0 & 10.0 & 8.0 & 56.8 & \textbf{58.0} & \textbf{100.0} & \textbf{100.0} \\
& AWM       & 10.0 & 10.0 & 19.3 & 24.0 & 10.0 & 8.0 & 22.5 & 20.0 & \textbf{100.0} & \textbf{100.0} \\
& Reflexion & 10.0 & 10.0 & 8.8 & 10.0 & 1.8 & 2.0 & 8.8 & 10.0 & 75.0 & \textbf{100.0} \\
& EvoTest   & 29.0 & 29.0 & 19.3 & 19.0 & 11.0 & 8.0 & \textbf{58.0} & \textbf{58.0} & \textbf{100.0} & \textbf{100.0} \\
& JitRL     & 60.0 & 60.0 & 60.0 & 60.0 & \textbf{33.5} & \textbf{32.0} & 36.0 & 53.0 & 76.5 & \textbf{100.0} \\
& \textbf{Ours}
& \textbf{75.0} & \textbf{83.0} & \textbf{60.5} & \textbf{74.0}
& \textbf{33.5} & \textbf{32.0} & \textbf{58.0} & \textbf{58.0}
& \textbf{100.0} & \textbf{100.0} \\
\bottomrule
\end{tabular}%
}

\end{table}

\begin{figure}[!t]
    \centering
    \includegraphics[width=\linewidth]{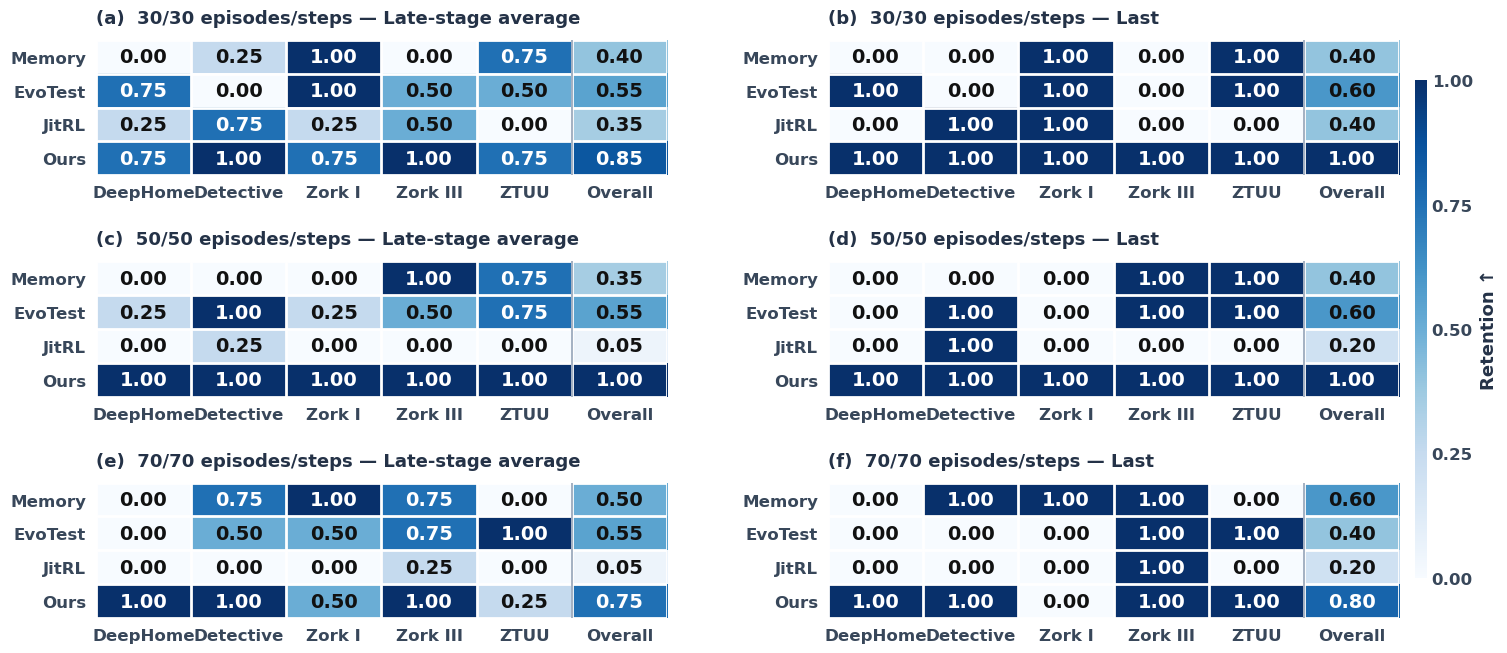}
\caption{Late-stage memory retention on Jericho ($\uparrow$). Darker indicates higher retention.}
    \label{fig:jericho-peak-success}
\end{figure}

\begin{table}[!t]
\captionsetup{font=small}
\centering
\begin{minipage}[t]{0.485\textwidth}
\vspace{0.1pt}
\centering
\footnotesize
\caption{Crafter achievement counts (0--22), with $100$ steps per episode. Last-4 denotes the final-four average; Last denotes the final episode's score ($\uparrow$).}
\vspace{1mm}
\label{tab:crafter-results}
\setlength{\tabcolsep}{3pt}
\renewcommand{\arraystretch}{1.1}
\begin{adjustbox}{max width=\linewidth}
\begin{tabular}{@{}l*{6}{r}@{}}
\toprule
\multirow{2}{*}{\textbf{Method}}
& \multicolumn{2}{c}{\textbf{10 Episodes}}
& \multicolumn{2}{c}{\textbf{20 Episodes}}
& \multicolumn{2}{c}{\textbf{30 Episodes}} \\
\cmidrule(lr){2-3}\cmidrule(lr){4-5}\cmidrule(lr){6-7}
& \textbf{Last-4} & \textbf{Last}
& \textbf{Last-4} & \textbf{Last}
& \textbf{Last-4} & \textbf{Last} \\
\midrule
Memory    & 7.0 & 7.0 & 8.3 & \textbf{9.0} & 4.0 & 5.0 \\
AWM       & 4.8 & 2.0 & 2.8 & 6.0 & 5.3 & \textbf{8.0} \\
Reflexion & 5.0 & 4.0 & 5.0 & 6.0 & 5.5 & 4.0 \\
EvoTest   & 6.5 & 7.0 & 6.0 & 7.0 & 6.0 & 6.0 \\
JitRL     & 5.8 & 6.0 & 6.8 & 7.0 & 6.0 & 7.0 \\
\textbf{Ours}
& \textbf{7.8} & \textbf{9.0}
& \textbf{9.0} & \textbf{9.0}
& \textbf{8.5} & \textbf{8.0} \\
\bottomrule
\end{tabular}
\end{adjustbox}
\end{minipage}\hfill
\begin{minipage}[t]{0.5\textwidth}
\vspace{0pt}
\centering
\footnotesize
\includegraphics[
    width=\linewidth,
    height=4.5cm,
keepaspectratio
]{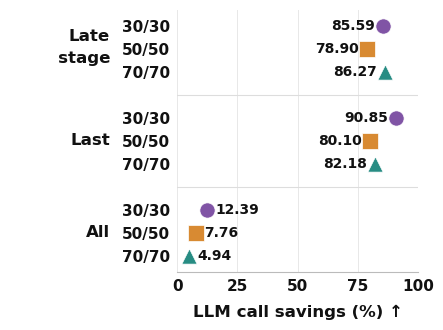}
\vspace{2mm}
\captionof{figure}{ Jericho candidate-generation call savings (\%, $\uparrow$)}
\label{fig:inference-savings}
\end{minipage}

\end{table}

\textbf{Performance Retention and Inference Savings (Q2).} When stored routes remain applicable, EARN combines reliable performance reproduction with fewer candidate-generation calls. At $50\times50$ on Jericho, Last-4 retention reaches 1.00 on every game, compared with five-game averages of 0.55 for EvoTest and 0.05 for JitRL (Figure~\ref{fig:jericho-peak-success}). EARN therefore matches its run-best scores throughout the final four episodes, rather than recovering only in the last episode. On Jericho, replay avoids 78.90--86.27\% of candidate-generation calls in the final four episodes. These final-window replay steps correspond to whole-run call savings of 4.94--12.39\%, averaged equally across games (Figure~\ref{fig:inference-savings}). Retention measures reproduction of each run's own peak, not global optimality. The saved calls could support additional exploration under a fixed inference budget, although the performance benefit of such reallocation remains untested.

\textbf{Experience Transfer (Q3).} In ScienceWorld, task variations change the environment across episodes, so stored routes may become inapplicable. EARN can still use retrieved experience and action-level evidence to adjust candidate preferences, while generating actions conditioned on the current observation. This separates experience reuse from fixed-route execution: changes in object locations or task requirements need not invalidate all previously acquired evidence. EARN achieves the best or tied Last-4 score in 14 of 15 task--budget settings (Table~\ref{tab:scienceworld-results}). EARN reuses experience without requiring an entire route to remain applicable: retrieved memory informs candidate generation, while supported action-level evidence adjusts preferences under the current observation. This enables selective reuse of prior experience even when environmental changes prevent direct replay.

\subsection{Ablation Study and Method Analysis}
\textbf{Ablation Studies. }We remove block-wise credit, step-wise credit, and conditional replay individually on ZTUU and Zork III to assess their contributions to final-episode performance and examine whether the gains come solely from replay. All settings use 30 episodes and 30 steps each. As shown in Table~\ref{tab:jericho-ablation}, removing either block-wise credit or conditional replay reduces Last-4 from 20.8 to 12.0 on ZTUU and from 4.0 to 3.0 on Zork III. The decline without block-wise credit, even with replay retained, supports contributions from both credit-based action adjustment and route reuse. The effect of step-wise credit varies across tasks: removing it hurts ZTUU, supporting the use of feedback from all trajectories to complement block-wise experience, but improves both Last-4 and Last on Zork III. Additional step-wise feedback therefore does not necessarily improve performance, motivating further analysis of its contribution strength.

\textbf{Sensitivity Analysis.} We vary the suppression coefficient $\eta\in\{0,0.5,0.85,1.0\}$ in Equation~\ref{eq:fusion}, keeping $\lambda_{\mathrm{step}}=0.35$ and the remaining settings fixed. As shown in Table~\ref{tab:suppression}, the default $\eta=0.85$ achieves the highest Last-4 score on ZTUU (20.8). On Zork III, $\eta=1.0$ achieves the highest Last-4 score (4.5), compared with 4.0 for the default; both attain a final-four minimum score of 4.0. On both tasks, $\eta=1.0$ outperforms $\eta=0$ in Last-4, although performance does not vary monotonically with suppression strength. Thus, retaining a stronger step-wise contribution does not necessarily improve performance. The default performs well on both tested tasks, ranking first on ZTUU and remaining close to the best tested setting on Zork III.

\begin{table}[!t]
\centering
\captionsetup{font=small}

\begin{minipage}[t]{0.47\textwidth}
\centering
\caption{Ablation results over 30 episodes and 30 steps each. Last-4 denotes the final-four-episode mean, and Last denotes the final-episode score ($\uparrow$). Bold indicates the highest score in each column.}
\vspace{1mm}
\label{tab:jericho-ablation}
\footnotesize
\setlength{\tabcolsep}{3pt}
\renewcommand{\arraystretch}{1.08}
\setlength{\aboverulesep}{2pt}
\setlength{\belowrulesep}{2pt}
\begin{adjustbox}{max width=\linewidth}
\begin{tabular}{@{}lrrrr@{}}
\toprule
\multirow{2}{*}{\textbf{Method}}
& \multicolumn{2}{c}{\textbf{ZTUU}}
& \multicolumn{2}{c}{\textbf{Zork III}} \\
\cmidrule(lr){2-3}\cmidrule(lr){4-5}
& \textbf{Last-4} & \textbf{Last}
& \textbf{Last-4} & \textbf{Last} \\
\midrule
Memory    & 3.8 & 5.0  & 1.8 & 2.0 \\
Reflexion & 0.0 & 0.0  & 0.8 & 1.0 \\
AWM       & 0.0 & 0.0  & 0.0 & 0.0 \\
EvoTest   & 2.5 & 5.0  & 2.3 & 2.0 \\
JitRL     & 6.3 & 10.0 & 2.5 & 2.0 \\
\midrule
\rowcolor{blue!6}
\textbf{EARN (Full)}
& \textbf{20.8} & \textbf{22.0} & 4.0 & 4.0 \\
w/o Block-wise Credit & 12.0 & 12.0 & 3.0 & 3.0 \\
w/o Step-wise Credit  & 16.5 & 17.0 & \textbf{5.0} & \textbf{5.0} \\
w/o Replay        & 12.0 & 12.0 & 3.0 & 3.0 \\
\bottomrule
\end{tabular}
\end{adjustbox}
\end{minipage}\hfill
\begin{minipage}[t]{0.51\textwidth}
\vspace{0pt}
\centering
\caption{Sensitivity to step-wise credit suppression on ZTUU and Zork III over 30 episodes with 30 steps each. The base step-wise credit weight is fixed at 0.35, while its effective contribution is attenuated by the suppression coefficient according to the strength of block-wise evidence. EARN (Full) uses the original suppression coefficient of 0.85. Min denotes the minimum scores over the final four episodes ($\uparrow$).}
\vspace{1mm}
\label{tab:suppression}
\footnotesize
\setlength{\tabcolsep}{3pt}
\renewcommand{\arraystretch}{1.08}
\setlength{\aboverulesep}{2pt}
\setlength{\belowrulesep}{2pt}
\begin{adjustbox}{max width=\linewidth}
\begin{tabular}{@{}lc*{6}{r}@{}}
\toprule
\multirow{2}{*}{\textbf{Method}}
& \multirow{2}{*}{\textbf{Supp.}}
& \multicolumn{3}{c}{\textbf{ZTUU}}
& \multicolumn{3}{c}{\textbf{Zork III}} \\
\cmidrule(lr){3-5}\cmidrule(lr){6-8}
& & \textbf{Last-4} & \textbf{Last} & \textbf{Min}
& \textbf{Last-4} & \textbf{Last} & \textbf{Min} \\
\midrule
\rowcolor{blue!6}
\textbf{Full} & 0.85
& \textbf{20.8} & \textbf{22.0} & \textbf{17.0}
& 4.0 & 4.0 & \textbf{4.0} \\
\midrule
\multirow{3}{*}{EARN}
& 0.0 & 16.0 & 16.0 & 16.0 & 3.0 & 3.0 & 3.0 \\
& 0.5 & 12.0 & 12.0 & 12.0 & 3.0 & 3.0 & 3.0 \\
& 1.0 & 16.8 & 18.0 & 13.0
& \textbf{4.5} & \textbf{5.0} & \textbf{4.0} \\
\bottomrule
\end{tabular}
\end{adjustbox}
\end{minipage}

\end{table}

\section{Conclusion}
We introduced \method, a training-free method that combines progress-relative block-wise credits with step-wise feedback to adjust candidate-action probabilities. Block-wise selection prioritizes high-performing trajectories, while step-wise updates retain feedback from every episode. Conditional route replay reuses applicable action sequences during the final episodes, bypassing LLM candidate generation and returning to memory-guided selection when replay ends. Across Jericho, ScienceWorld, and Crafter, \method achieves the best or tied-best Last-4 score in 31 of 33 task--budget settings, with substantial candidate-generation call savings on Jericho. Future work will examine experience reuse under changing conditions and stochastic transitions.

\bibliographystyle{unsrtnat}
\bibliography{references}

@inproceedings{
shinn2023reflexion,
title={Reflexion: language agents with verbal reinforcement learning},
author={Noah Shinn and Federico Cassano and Ashwin Gopinath and Karthik R Narasimhan and Shunyu Yao},
booktitle={Thirty-seventh Conference on Neural Information Processing Systems},
year={2023},
url={https://openreview.net/forum?id=vAElhFcKW6}
}

@inproceedings{
he2026evotest,
title={EvoTest: Evolutionary Test-Time Learning for Self-Improving Agentic Systems},
author={Yufei He and Juncheng Liu and Yue Liu and Yibo Li and Tri Cao and Zhiyuan Hu and Xinxing Xu and Bryan Hooi},
booktitle={The Fourteenth International Conference on Learning Representations},
year={2026},
url={https://openreview.net/forum?id=JFnnajbkvP}
}

@inproceedings{
li2026justintime,
title={Just-In-Time Reinforcement Learning: Continual Learning in {LLM} Agents Without Gradient Updates},
author={Yibo Li and Zijie Lin and Ailin Deng and Xuan Zhang and Yufei He and Shuo Ji and Tri Cao and Bryan Hooi},
booktitle={Proceedings of the 43rd International Conference on Machine Learning},
series={Proceedings of Machine Learning Research},
volume={306},
year={2026},
url={https://openreview.net/forum?id=pLvye0zHUC}
}

@article{
wang2024voyager,
title={Voyager: An Open-Ended Embodied Agent with Large Language Models},
author={Guanzhi Wang and Yuqi Xie and Yunfan Jiang and Ajay Mandlekar and Chaowei Xiao and Yuke Zhu and Linxi Fan and Anima Anandkumar},
journal={Transactions on Machine Learning Research},
issn={2835-8856},
year={2024},
url={https://openreview.net/forum?id=ehfRiF0R3a},
note={}
}

@inproceedings{zhang2023rememberer,
 author = {Zhang, Danyang and Chen, Lu and Zhang, Situo and Xu, Hongshen and Zhao, Zihan and Yu, Kai},
 booktitle = {Advances in Neural Information Processing Systems},
 doi = {10.52202/075280-3419},
 pages = {78227--78239},
 publisher = {Curran Associates, Inc.},
 title = {Large Language Models Are Semi-Parametric Reinforcement Learning Agents},
 volume = {36},
 year = {2023}
}

@inproceedings{majumder2024clin,
 title = {{CLIN}: A Continually Learning Language Agent for Rapid Task Adaptation and Generalization},
 author = {Bodhisattwa Prasad Majumder and Bhavana Dalvi Mishra and Peter Jansen and Oyvind Tafjord and Niket Tandon and Li Zhang and Chris Callison-Burch and Peter Clark},
 booktitle = {First Conference on Language Modeling},
 year = {2024},
 url = {https://openreview.net/forum?id=xS6zx1aBI9}
}

@article{zhao2024expel,
 title = {{ExpeL}: {LLM} Agents Are Experiential Learners},
 author = {Andrew Zhao and Daniel Huang and Quentin Xu and Matthieu Lin and Yong-Jin Liu and Gao Huang},
 journal = {Proceedings of the AAAI Conference on Artificial Intelligence},
 year = {2024},
 volume = {38},
 number = {17},
 pages = {19632--19642},
 doi = {10.1609/aaai.v38i17.29936},
 url = {https://ojs.aaai.org/index.php/AAAI/article/view/29936}
}

@inproceedings{zheng2024synapse,
 author = {Zheng, Longtao and Wang, Rundong and Wang, Xinrun and An, Bo},
 booktitle = {International Conference on Learning Representations},
 pages = {19036--19066},
 title = {Synapse: Trajectory-as-Exemplar Prompting with Memory for Computer Control},
 volume = {2024},
 year = {2024}
}

@inproceedings{wang2025awm,
 title = {Agent Workflow Memory},
 author = {Wang, Zora Zhiruo and Mao, Jiayuan and Fried, Daniel and Neubig, Graham},
 booktitle = {Proceedings of the 42nd International Conference on Machine Learning},
 pages = {63897--63911},
 year = {2025},
 volume = {267},
 series = {Proceedings of Machine Learning Research},
 publisher = {PMLR},
 url = {https://proceedings.mlr.press/v267/wang25bx.html}
}

@inproceedings{ouyang2026reasoningbank,
 title = {ReasoningBank: Scaling Agent Self-Evolving with Reasoning Memory},
 author = {Siru Ouyang and Jun Yan and I-Hung Hsu and Yanfei Chen and Ke Jiang and Zifeng Wang and Rujun Han and Long T. Le and Samira Daruki and Xiangru Tang and Vishy Tirumalashetty and George Lee and Mahsan Rofouei and Hangfei Lin and Jiawei Han and Chen-Yu Lee and Tomas Pfister},
 booktitle = {The Fourteenth International Conference on Learning Representations},
 year = {2026},
 url = {https://openreview.net/forum?id=jL7fwchScm}
}

@inproceedings{sarukkai2025selfgenerated,
 author = {Sarukkai, Vishnu and Xie, Zhiqiang and Fatahalian, Kayvon},
 booktitle = {Advances in Neural Information Processing Systems},
 doi = {10.52202/085713-2158},
 pages = {64392--64425},
 publisher = {Curran Associates, Inc.},
 title = {Self-Generated In-Context Examples Improve {LLM} Agents for Sequential Decision-Making Tasks},
 volume = {38},
 year = {2025}
}

@inproceedings{fu2024autoguide,
 author = {Fu, Yao and Kim, Dong-Ki and Kim, Jaekyeom and Sohn, Sungryull and Logeswaran, Lajanugen and Bae, Kyunghoon and Lee, Honglak},
 booktitle = {Advances in Neural Information Processing Systems},
 doi = {10.52202/079017-3811},
 pages = {119919--119948},
 publisher = {Curran Associates, Inc.},
 title = {AutoGuide: Automated Generation and Selection of Context-Aware Guidelines for Large Language Model Agents},
 volume = {37},
 year = {2024}
}

@misc{zheng2025lifelongagentbenchevaluatingllmagents,
      title={LifelongAgentBench: Evaluating LLM Agents as Lifelong Learners}, 
      author={Junhao Zheng and Xidi Cai and Qiuke Li and Duzhen Zhang and ZhongZhi Li and Yingying Zhang and Le Song and Qianli Ma},
      year={2025},
      eprint={2505.11942},
      archivePrefix={arXiv},
      primaryClass={cs.AI},
      url={https://arxiv.org/abs/2505.11942}, 
}

@inproceedings{NEURIPS2019_16105fb9,
 author = {Arjona-Medina, Jose A. and Gillhofer, Michael and Widrich, Michael and Unterthiner, Thomas and Brandstetter, Johannes and Hochreiter, Sepp},
 booktitle = {Advances in Neural Information Processing Systems},
 editor = {H. Wallach and H. Larochelle and A. Beygelzimer and F. d\textquotesingle Alch\'{e}-Buc and E. Fox and R. Garnett},
 pages = {},
 publisher = {Curran Associates, Inc.},
 title = {RUDDER: Return Decomposition for Delayed Rewards},
 url = {https://proceedings.neurips.cc/paper_files/paper/2019/file/16105fb9cc614fc29e1bda00dab60d41-Paper.pdf},
 volume = {32},
 year = {2019}
}

@inproceedings{
toneva2018an,
title={An Empirical Study of Example Forgetting during Deep Neural Network Learning},
author={Mariya Toneva and Alessandro Sordoni and Remi Tachet des Combes and Adam Trischler and Yoshua Bengio and Geoffrey J. Gordon},
booktitle={International Conference on Learning Representations},
year={2019},
url={https://openreview.net/forum?id=BJlxm30cKm},
}

@inproceedings{
tirumala2022memorization,
title={Memorization Without Overfitting:  Analyzing the Training Dynamics of Large Language Models},
author={Kushal Tirumala and Aram H. Markosyan and Luke Zettlemoyer and Armen Aghajanyan},
booktitle={Advances in Neural Information Processing Systems},
editor={Alice H. Oh and Alekh Agarwal and Danielle Belgrave and Kyunghyun Cho},
year={2022},
url={https://openreview.net/forum?id=u3vEuRr08MT}
}

@inproceedings{NEURIPS2019_fa7cdfad,
 author = {Rolnick, David and Ahuja, Arun and Schwarz, Jonathan and Lillicrap, Timothy and Wayne, Gregory},
 booktitle = {Advances in Neural Information Processing Systems},
 editor = {H. Wallach and H. Larochelle and A. Beygelzimer and F. d\textquotesingle Alch\'{e}-Buc and E. Fox and R. Garnett},
 pages = {},
 publisher = {Curran Associates, Inc.},
 title = {Experience Replay for Continual Learning},
 url = {https://proceedings.neurips.cc/paper_files/paper/2019/file/fa7cdfad1a5aaf8370ebeda47a1ff1c3-Paper.pdf},
 volume = {32},
 year = {2019}
}

@inproceedings{kang2025your,
  title={Do your best and get enough rest for continual learning},
  author={Kang, Hankyul and Seifer, Gregor and Lee, Donghyun and Ryu, Jongbin},
  booktitle={2025 IEEE/CVF Conference on Computer Vision and Pattern Recognition (CVPR)},
  pages={10077--10086},
  year={2025},
  organization={IEEE}
}

@inproceedings{
wang2023selfconsistency,
title={Self-Consistency Improves Chain of Thought Reasoning in Language Models},
author={Xuezhi Wang and Jason Wei and Dale Schuurmans and Quoc V Le and Ed H. Chi and Sharan Narang and Aakanksha Chowdhery and Denny Zhou},
booktitle={The Eleventh International Conference on Learning Representations },
year={2023},
url={https://openreview.net/forum?id=1PL1NIMMrw}
}

@inproceedings{hausknecht2020interactive,
  title={Interactive fiction games: A colossal adventure},
  author={Hausknecht, Matthew and Ammanabrolu, Prithviraj and C{\^o}t{\'e}, Marc-Alexandre and Yuan, Xingdi},
  booktitle={Proceedings of the AAAI Conference on Artificial Intelligence},
  volume={34},
  number={05},
  pages={7903--7910},
  year={2020}
}

@inproceedings{wang2022scienceworld,
  title={Scienceworld: Is your agent smarter than a 5th grader?},
  author={Wang, Ruoyao and Jansen, Peter and C{\^o}t{\'e}, Marc-Alexandre and Ammanabrolu, Prithviraj},
  booktitle={Proceedings of the 2022 Conference on Empirical Methods in Natural Language Processing},
  pages={11279--11298},
  year={2022}
}

@misc{gemmateam2026gemma4technicalreport,
      title={Gemma 4 Technical Report}, 
      author={Gemma Team and Sherif El Abd and Vaibhav Aggarwal and Robin Algayres and Alek Andreev and Olivier Bachem and Ian Ballantyne and Cormac Brick and Victor Cărbune and Michelle Casbon and Mayank Chaturvedi and Aditya Chawla and Victor Cotruta and Alice Coucke and Phil Culliton and Robert Dadashi and Lucas Dixon and Mohamed Elhawaty and Utku Evci and Clément Farabet and Johan Ferret and Filippo Galgani and Sertan Girgin and Jean-Bastien Grill and Maarten Grootendorst and Jiaxian Guo and Cassidy Hardin and Yanzhang He and Steven M. Hernandez and Omri Homburger and Léonard Hussenot and Juyeong Ji and Armand Joulin and Aishwarya Kamath and Parnian Kassraie and Olivier Lacombe and Preethi Lahoti and Gaël Liu and Gus Martins and Luciano Martins and Tatiana Matejovicova and Ramona Merhej and Nikola Momchev and Sneha Mondal and Ryan Mullins and Sindhu Raghuram Panyam and Shreya Pathak and Sarah Perrin and André Susano Pinto and Etienne Pot and Angéline Pouget and Alexandre Ramé and Sabela Ramos and Douglas Reid and David Rim and Morgane Rivière and Karsten Roth and Louis Rouillard and Omar Sanseviero and Pier Giuseppe Sessa and Shane Settle and Danila Sinopalnikov and Sara Smoot and Piotr Stanczyk and Andreas Steiner and Lawrence Stewart and Ilya Tolstikhin and Michael Tschannen and Anton Tsitsulin and Nino Vieillard and Renjie Wu and Pingmei Xu and Haichuan Yang and Edouard Yvinec and Biao Zhang and Li Zhang and Joe Zou and Nicolas Aagnes and Abdelrahman Abdelhamed and Jakub Adamek and Shivani Agrawal and Shubham Agrawal and Ibrahim Alabdulmohsin and Jean Baptiste Alayrac and Uri Alon and Chandramouli Amarnath and Ankesh Anand and Chrysovalantis Anastasiou and Setareh Ariafar and François-Xavier Aubet and Kyriakos Axiotis and Federico Barbero and Joelle Barral and Alexei Bendebury and Urs Bergmann and Stanley Bileschi and Kat Black and Mathieu Blondel and Sebastian Borgeaud and Arthur Bražinskas and Ryan Burnell and Robert Busa-Fekete and Mu Cai and Daniele Calandriello and Glenn Cameron and Charlotte Caucheteux and Rahma Chaabouni and Garima Chadha and Jetha Chan and Blake Jianhang Chen and Jesse Chen and Lin Chen and Xu Chen and Derek Cheng and Tzu-hsiang Chien and Nikolai Chinaev and Yi Chou and Zhaohui Chu and Benjamin Coleman and Pooja Consul and Sam Conway-Rahman and Scott Crowell and Dylan Cutler and Vivek Dani and Samira Daruki and Anil Das and Daniel Deutsch and Nishanth Dikkala and Li Ding and Qiuhan Ding and Shenil Dodhia and Konstantin Donhauser and Tulsee Doshi and Anca Dragan and Alex Druinsky and Sahil Dua and Zoltan Egyed and Danielle Eisenbud and Daniel Eppens and Cindy Fan and Bahare Fatemi and Yassir Fathullah and Vlad Feinberg and Milen Ferev and Sebastian Flennerhag and Takumi Fujimoto and João Gabriel Oliveira and Isaac Galatzer-Levy and João Gante and Simon Geisler and Soham Ghosal and Antonious M. Girgis and Tamara von Glehn and Alec Go and Alhaad Gokhale and Alex Grills and Yiming Gu and Mayank Gupta and Pramod Gupta and Guru Guruganesh and Raia Hadsell and Hamza Harkous and Jitendra Harlalka and Demis Hassabis and Anja Hauth and Joe Heyward and Arian Hosseini and Chih-Yang Hsia and I-Hung Hsu and Xiaopeng Huang and Yangsibo Huang and Kevin Hui and Adrian Hutter and Te I and Fotis Iliopoulos and Advait Jain and Ganesh Jawahar and Ziwei Ji and Qilin Jin and Melvin Johnson and Kandarp Joshi and Arun Kandoor and Wang-Cheng Kang and Koray Kavukcuoglu and Mehran Kazemi and Kathleen Kenealy and Amr Khalifa and Phoebe Kirk and Ivan Korotkov and Suraj Kothawade and Vitaly Kovalev and Neel Kovelamudi and Adam Kraft and Ravin Kumar and Vivek Kumar and Harish Kuppam and Justin Lannin and Chen-Yu Lee and Seungji Lee and Dmitry Lepikhin and Alon Levkovitch and Dongdong Li and Qiujia Li and Valentin Liévin and Ethan Lin and Ziqian Lin and Casper Liu and Tianlin Liu and Tianqi Liu and Xin Liu and Ivan Lobov and Mayank Lunayach and Min Ma and Gagan Madan and Andrii Maksai and Eric Malmi and Michal Matuszak and Daniel McDuff and Gaurav Menghani and Maciej Mikuła and Daniil Mirylenka and Karolis Misiunas and Vedant Misra and Andreea Mitran and Kareem Mohamed and Maksim Mukha and Eric Noland and James O'Donnell and Brendan O'Donoghue and Kate Olszewska and Bernett Orlando and Wanqiong Pan and Rina Panigrahy and Unnati Parekh and Nicolas Perez-Nieves and Chunjong Park and Eric Paskie and Liqian Peng and Bryce Petrini and Slav Petrov and Jonas Pfeiffer and Bilal Piot and Martyna Plomecka and Siim Poder and Octavio Ponce and Arijit Pramanik and David Racz and Anish Rajan and Michelle Ramanovich and Anand Rao and Marvin Ritter and Vitor Rodrigues and Evan Rosen and Mikołaj Rybiński and Noveen Sachdeva and Michaël E. Sander and Rohit Sathyanarayana and Sagar Savla and Samuel Schmidgall and Tal Schuster and George Scrivener and Benoit Seguin and Andrew Sellergren and Aliaksei Severyn and Izhak Shafran and Dhruv Shah and Bobak Shahriari and Yuan Shangguan and Ashish Shenoy and Pradeep Shenoy and Rakesh Shivanna and Pauline Sho and Lucas Spangher and Wojciech Stokowiec and Tim Strother and Yao Su and Yinghao Sun and Mukund Sundararajan and Andrea Tacchetti and Mor Hazan Taege and Pouya Tafti and Jean Tarbouriech and Chetan Tekur and Shantanu Thakoor and Rahul Thapa and Madeleine Traverse and Lenart Treven and Tao Tu and Chien Te Tung and Çağlar Ünlü and Petar Veličković and Malini Pooni Venkat and Sagar Gubbi Venkatesh and Vidya Venkiteswaran and Francesco Visin and Alex Vitvitskyi and Kiran Vodrahalli and Weiyi Wang and Xin Wang and Tris Warkentin and Jan Wassenberg and John Wieting and Cindy Wu and Lechao Xiao and Hao Xu and Yuhui Xu and Fuzhao Xue and Arun Yadav and Jun Yan and Antoine Yang and Lin Yang and Ming-Hsuan Yang and Ziyu Ying and Jae Hyeon Yoo and Morteza Zadimoghaddam and Sajjad Zafar and Fred Zhang and Jiageng Zhang and Jianyi Zhang and Xiaofan Zhang and Chao Zhao and David Zhou and Chen Zou},
      year={2026},
      eprint={2607.02770},
      archivePrefix={arXiv},
      primaryClass={cs.CL},
      url={https://arxiv.org/abs/2607.02770}, 
}

@misc{kwon2023efficientmemorymanagementlarge,
      title={Efficient Memory Management for Large Language Model Serving with PagedAttention}, 
      author={Woosuk Kwon and Zhuohan Li and Siyuan Zhuang and Ying Sheng and Lianmin Zheng and Cody Hao Yu and Joseph E. Gonzalez and Hao Zhang and Ion Stoica},
      year={2023},
      eprint={2309.06180},
      archivePrefix={arXiv},
      primaryClass={cs.LG},
      url={https://arxiv.org/abs/2309.06180}, 
}

@misc{wang2020minilmdeepselfattentiondistillation,
      title={MiniLM: Deep Self-Attention Distillation for Task-Agnostic Compression of Pre-Trained Transformers}, 
      author={Wenhui Wang and Furu Wei and Li Dong and Hangbo Bao and Nan Yang and Ming Zhou},
      year={2020},
      eprint={2002.10957},
      archivePrefix={arXiv},
      primaryClass={cs.CL},
      url={https://arxiv.org/abs/2002.10957}, 
}

@inproceedings{
    hafner2022benchmarking,
    title={Benchmarking the Spectrum of Agent Capabilities},
    author={Danijar Hafner},
    booktitle={International Conference on Learning Representations},
    year={2022},
    url={https://openreview.net/forum?id=1W0z96MFEoH}
}

@misc{yao2023reactsynergizingreasoningacting,
      title={ReAct: Synergizing Reasoning and Acting in Language Models}, 
      author={Shunyu Yao and Jeffrey Zhao and Dian Yu and Nan Du and Izhak Shafran and Karthik Narasimhan and Yuan Cao},
      year={2023},
      eprint={2210.03629},
      archivePrefix={arXiv},
      primaryClass={cs.CL},
      url={https://arxiv.org/abs/2210.03629}, 
}

@article{Kirkpatrick_2017,
   title={Overcoming catastrophic forgetting in neural networks},
   volume={114},
   ISSN={1091-6490},
   url={http://dx.doi.org/10.1073/pnas.1611835114},
   DOI={10.1073/pnas.1611835114},
   number={13},
   journal={Proceedings of the National Academy of Sciences},
   publisher={National Academy of Sciences},
   author={Kirkpatrick, James and Pascanu, Razvan and Rabinowitz, Neil and Veness, Joel and Desjardins, Guillaume and Rusu, Andrei A. and Milan, Kieran and Quan, John and Ramalho, Tiago and Grabska-Barwinska, Agnieszka and Hassabis, Demis and Clopath, Claudia and Kumaran, Dharshan and Hadsell, Raia},
   year={2017},
   month=Mar, pages={3521–3526} }

@misc{lopezpaz2022gradientepisodicmemorycontinual,
      title={Gradient Episodic Memory for Continual Learning}, 
      author={David Lopez-Paz and Marc'Aurelio Ranzato},
      year={2022},
      eprint={1706.08840},
      archivePrefix={arXiv},
      primaryClass={cs.LG},
      url={https://arxiv.org/abs/1706.08840}, 
}

@misc{chaudhry2019tinyepisodicmemoriescontinual,
      title={On Tiny Episodic Memories in Continual Learning}, 
      author={Arslan Chaudhry and Marcus Rohrbach and Mohamed Elhoseiny and Thalaiyasingam Ajanthan and Puneet K. Dokania and Philip H. S. Torr and Marc'Aurelio Ranzato},
      year={2019},
      eprint={1902.10486},
      archivePrefix={arXiv},
      primaryClass={cs.LG},
      url={https://arxiv.org/abs/1902.10486}, 
}

@misc{aljundi2019onlinecontinuallearningmaximally,
      title={Online Continual Learning with Maximally Interfered Retrieval}, 
      author={Rahaf Aljundi and Lucas Caccia and Eugene Belilovsky and Massimo Caccia and Min Lin and Laurent Charlin and Tinne Tuytelaars},
      year={2019},
      eprint={1908.04742},
      archivePrefix={arXiv},
      primaryClass={cs.LG},
      url={https://arxiv.org/abs/1908.04742}, 
}

@misc{buzzega2020darkexperiencegeneralcontinual,
      title={Dark Experience for General Continual Learning: a Strong, Simple Baseline}, 
      author={Pietro Buzzega and Matteo Boschini and Angelo Porrello and Davide Abati and Simone Calderara},
      year={2020},
      eprint={2004.07211},
      archivePrefix={arXiv},
      primaryClass={stat.ML},
      url={https://arxiv.org/abs/2004.07211}, 
}

@misc{wang2022improvingtaskfreecontinuallearning,
      title={Improving Task-free Continual Learning by Distributionally Robust Memory Evolution}, 
      author={Zhenyi Wang and Li Shen and Le Fang and Qiuling Suo and Tiehang Duan and Mingchen Gao},
      year={2022},
      eprint={2207.07256},
      archivePrefix={arXiv},
      primaryClass={cs.LG},
      url={https://arxiv.org/abs/2207.07256}, 
}

\clearpage
\appendix
\section{Main notation}
\label{app:notation}
\begin{table}[htbp]
\centering
\footnotesize
\caption{Main notation used in \method.}
\vspace{1mm}
\label{tab:notation}
\renewcommand{\arraystretch}{1.15}
\setlength{\tabcolsep}{6pt}
\begin{tabular}{@{}p{0.28\linewidth}p{\dimexpr0.72\linewidth-12pt\relax}@{}}
\hline
\textbf{Notation} & \textbf{Description} \\
\hline
\multicolumn{2}{@{}l}{\textit{Problem Formulation}} \\[5pt]
$K,\ T_k,\ W$ & Episode budget, steps in episode $k$, and number of final evaluation episodes. \\
$s_{k,t},\ a_{k,t},\ y_{k,t}$ & Execution context, action, and environment feedback at step $t$ of episode $k$. \\
$\theta,\ x_{k,t}$ & Frozen LLM parameters and the prompt at step $t$ of episode $k$. \\
$S_{k,t},\ S_k$ & Total score after step $t$ and episode score assigned by the scoring criterion. \\
$J_{\mathrm{term}}$ & Expected average episode score over the final $W$ episodes. \\[6pt]
\multicolumn{2}{@{}l}{\textit{Credit and Memory}} \\[5pt]
$\tau_k$ & Episode trajectory: execution contexts, actions, and feedback. \\
$B,\ F_b$ & Block size and progress frontier fixed at the start of block $b$. \\
$e,\ S_e$ & Scoring event and total score immediately after that event. \\
$c_{e,i}$ & Block-wise credit assigned to action $i$ in the segment for scoring event $e$. \\
$\tilde r_{k,t},\ C^{\mathrm{step}}_{k,t}$ & Constructed numerical signal and step-wise credit. \\
$\M_k,\ \mathcal R_k^\star,\ \mathcal E_k$ & Persistent memory, stored route, and action-credit records. \\[6pt]
\multicolumn{2}{@{}l}{\textit{Probability Adjustment and Replay}} \\[5pt]
$Q_{\mathrm{blk}},\ Q_{\mathrm{step}}$ & Block-wise and step-wise credit estimates. \\
$V_{\mathrm{blk}},\ V_{\mathrm{step}}$ & Means subtracted to obtain the centered credit estimates. \\
$A_{\mathrm{blk}},\ A_{\mathrm{step}},\ A_{\mathcal M}$ & Centered block-wise and step-wise credit estimates, and their combined credit estimate. \\
$n_{\mathrm{blk}},\ r_{\mathrm{blk}}$ & Number of matching block-wise credit records and the resulting scaling factor. \\
$g_{\mathrm{blk}},\ g_{\mathrm{step}},\ q_{\mathrm{step}}$ & Gates based on matching record counts and the largest scaling factor among matching step-wise records. \\
$p^0,\ p_{\mathcal M}$ & Base and adjusted distributions over candidate actions. \\
$z_k,\ \chi_{k,t}$ & Replay-episode indicator and replay-check indicator. \\
\hline
\end{tabular}
\end{table}

\section{Additional Related Work}
\paragraph{Attribution and Credit Assignment in Sequential Decision Making} Credit assignment in sequential decision making aims to determine which historical actions led to delayed outcomes. RUDDER reallocates delayed rewards by decomposing returns, attributing them back to critical decision points; similarly, JitRL estimates the advantage of current candidate actions using returns from past trajectories \citep{NEURIPS2019_16105fb9,li2026justintime}.

\paragraph{Memory Retention} Prior work studies example-level forgetting \citep{toneva2018an} and memorization dynamics in language models \citep{tirumala2022memorization}. Continual learning methods mitigate forgetting through parameter regularization \citep{Kirkpatrick_2017}, episodic-memory constraints \citep{lopezpaz2022gradientepisodicmemorycontinual}, and replay with small memory buffers \citep{chaudhry2019tinyepisodicmemoriescontinual}. Experience Replay for Continual Learning \citep{NEURIPS2019_fa7cdfad} uses replay to reduce forgetting, while subsequent work explores interference-based replay selection \citep{aljundi2019onlinecontinuallearningmaximally} and preservation of past model outputs \citep{buzzega2020darkexperiencegeneralcontinual}. Do Your Best and Get Enough Rest for Continual Learning \citep{kang2025your} examines how replay intervals affect retention. Distributionally robust memory evolution dynamically adjusts the memory data distribution to reduce buffer overfitting and alleviate forgetting \citep{wang2022improvingtaskfreecontinuallearning}. In our setting with a frozen model, we define late-stage memory retention as the fraction of final episodes attaining the run's peak score.

\paragraph{Continual Learning for Agentic Systems} Continual learning requires an agent to retain previously acquired effective behaviors while assimilating new experiences. Voyager achieves long-term capability accumulation in open worlds through automatic curriculum design and an executable skills library, whereas studies like LifelongAgentBench highlight that simply saving and replaying all historical experiences are prone to interference from irrelevant information and context length limitations \citep{wang2024voyager,zheng2025lifelongagentbenchevaluatingllmagents}. Our study addresses cross-episode adaptation under a fixed budget: the environment resets between episodes, but the memory persists.

\section{Additional Method Details}
\label{sec:method_details}
\subsection{Benchmarks and Tasks}

We evaluate on three interactive benchmarks: Jericho, ScienceWorld, and Crafter. The selected environments cover text-based adventure games, scientific tasks, and visual survival gameplay.

\paragraph{Jericho.} Jericho \citep{hausknecht2020interactive} provides text-based interactive fiction environments. Agents read observations and issue commands to navigate locations, inspect objects, and interact with the game world. We use five games:
\begin{itemize}
\item \textbf{DeepHome.} The player explores an abandoned mountain kingdom and prepares it for its inhabitants' return by restoring essential systems, opening gates, and addressing threats.

\item \textbf{Detective.} The player investigates the mayor's murder by visiting locations, examining information, collecting objects, and navigating potentially dangerous encounters.

\item \textbf{Zork I.} The player explores the Great Underground Empire, solves puzzles, and collects treasures to return to the trophy case in the white house.

\item \textbf{Zork III.} The player completes trials in the Great Underground Empire to demonstrate worthiness to succeed the Dungeon Master. Progress involves navigation, object use, and interactions with characters.

\item \textbf{ZTUU.} In Zork: The Undiscovered Underground, the player explores a newly discovered underground region and seeks to emerge successfully by navigating rooms and solving object-based puzzles.
\end{itemize}

\paragraph{ScienceWorld.} ScienceWorld \citep{wang2022scienceworld} is a text-based simulation environment for elementary science tasks. Agents manipulate objects, use instruments, and perform experiments to satisfy task-specific goals. We evaluate five tasks, each with multiple variations:
\begin{itemize}
\item \textbf{Conductivity.} Test whether a specified material conducts electricity by constructing a circuit and observing its behavior.

\item \textbf{Unknown Conductivity.} Determine the electrical conductivity of an unknown substance through circuit-based testing.

\item \textbf{Plant.} Grow a specified plant by preparing suitable growing conditions and providing the resources needed for its development.

\item \textbf{Power.} Power an electrical component using an energy source that satisfies the requested renewable or nonrenewable category.

\item \textbf{Melting Known.} Measure the melting point of a specified known substance using heating equipment and temperature measurements.
\end{itemize}

\paragraph{Crafter.} Crafter \citep{hafner2022benchmarking} is a visual, open-world survival game with procedurally generated environments. The agent gathers resources, crafts tools, builds useful structures, and manages survival needs. Progress is measured through 22 achievements, ranging from collecting basic resources to producing advanced tools. These achievements have dependencies, requiring the agent to carry out sequences of preparatory actions. We use a multimodal interface combining visual observations with textual information.

\paragraph{Memory storage.} Alongside the stored route and action-credit records, persistent memory retains aggregate credit statistics and trajectories awaiting block-wise selection.

\subsection{Block Size and Progress Frontier}
\label{app:block_frontier}

\paragraph{Block size.} We use blocks of $B=5$ acquisition episodes and select the best-performing trajectory within each block for milestone credit attribution. This provides several trajectories for comparison while keeping the delay before block-wise memory updates short. Smaller blocks offer fewer alternatives, whereas larger blocks delay the use of selected experience. Step-wise credit records continue to update after every episode.

This choice is inspired by self-consistency: \citet{wang2023selfconsistency} suggest starting with five or ten reasoning paths to balance performance and computational cost. Unlike their answer aggregation procedure, EARN selects among sequential interaction trajectories using environment feedback. We therefore adopt $B=5$ as a practical setting.

\paragraph{Progress Frontier.}
\label{app:frontier}
Let $P_k$ denote the route score from episode $k$ used to update the progress frontier. We locate the first action that reaches the episode's highest total score. If this action terminates the environment, $P_k$ is the score immediately before it; otherwise, $P_k$ is the score immediately after it. This rule also applies to successful termination.

After episode $k$, we consider the most recent $W_F=5$ episodes, including episode $k$, indexed by $\mathcal W_k=\{\max(1,k-W_F+1),\ldots,k\}$. We round each $P_j$ to six decimal places before comparing values. The candidate frontier is the highest positive route score recorded in at least two episodes within this window:
\begin{equation}
\widehat F_k =
\max\left(
\{0\}\cup
\left\{
r>0:
\sum_{j\in\mathcal W_k}\mathbf{1}[P_j=r]\ge 2
\right\}
\right).
\end{equation}
Here, $\mathbf{1}[\cdot]$ is the indicator function. Each episode contributes one route score, rather than all intermediate scores along its trajectory.

Let $F^{\mathrm{sys}}_k$ denote the progress frontier after episode $k$. We initialize it to zero and update it without allowing decreases:
\begin{equation}
F^{\mathrm{sys}}_0=0,\qquad
F^{\mathrm{sys}}_k
=\max\left(F^{\mathrm{sys}}_{k-1},\widehat F_k\right).
\end{equation}
If no positive route score appears at least twice in the window, $\widehat F_k=0$ and the frontier remains unchanged.

Let $k_b$ denote the first episode of block $b$. We fix the block's frontier before that episode:
\begin{equation}
F_b=F^{\mathrm{sys}}_{k_b-1}.
\end{equation}
All block-wise credit attribution in block $b$ uses this fixed value, even if the system frontier changes during the block. When $F_b=0$, the new-progress segment starts at the beginning of the trajectory, subject to the segment-length limit.

\paragraph{Example of Frontier Updates.} Suppose a block has just ended, the frontier before the latest update is 10, and the five most recent route scores are $(10,12,10,15,12)$. The candidate frontier is $\widehat F_k=12$: both 10 and 12 appear twice, whereas 15 appears only once. Thus, $F^{\mathrm{sys}}_k=\max(10,12)=12$. If the next episode starts a new block, that block uses $F_b=12$. Score-increase events ending at or below 12 receive reduced credit, while those ending above 12 receive higher credit. The block's reference remains fixed at 12 throughout the block.

\paragraph{Trajectory selection and credited segments.} After comparing the highest total scores, trajectory selection breaks ties by the score of the replayable prefix, fewer steps to the highest score, absence of terminal failure at that point, fewer local cycles, and earlier episode, in that order. Let $t_e$ be the step of the $e$th score increase in the selected trajectory, so $S_e=S_{k,t_e}$. If $S_e\leq F_b$, the credited segment begins after the preceding score increase. If $S_e>F_b$, the segment begins after the trajectory first reaches or exceeds $F_b$; when the score increase at $t_e$ first crosses $F_b$, the segment instead begins after the preceding score increase. Each segment ends at $t_e$ and retains only the most recent actions if its length exceeds the configured limit. We use $\rho=1/2$ in Equation~\ref{eq:selected_credit}. We limit repeated admission of identical action segments.

\paragraph{Notation for Block-wise Credit Attribution.} The symbols in the credit attribution formula are defined as follows:
\begin{itemize}
\item $e$: A scoring event in the selected trajectory, corresponding to a positive increase in the environment score. A trajectory may contain multiple scoring events.

\item $L_e$: The number of actions in the attribution segment associated with event $e$.

\item $i$: The action's position within this segment, indexed from $1$ to $L_e$. The final position $i=L_e$ corresponds to the scoring action.

\item $\ell_e$: The initial reference value for interpolation, corresponding to relative position $i/L_e=0$. It equals $0$ for repeated progress and $\rho$ for new progress.

\item $h_e$: The terminal interpolation value, equal to the scoring action's credit. It equals $\alpha_{\mathrm{pre}}$ for repeated progress and $1$ for new progress.

\item $c_{e,i}$: The credit assigned to the $i$-th action in the segment for event $e$.
\end{itemize}
Since $i$ starts at $1$, the first action receives $\ell_e+(h_e-\ell_e)/L_e$, rather than $\ell_e$. The final action receives $h_e$.

\subsection{Block-wise Credit Estimates}
\label{app:block_value}

For execution context $s$ and candidate action $a$, let $\{c_j\}_{j=1}^{n_{\mathrm{blk}}}$ denote the block-wise credits in records matching both $s$ and $a$ exactly. For $n_{\mathrm{blk}}>0$, their mean $\mu_{\mathrm{blk}}$ and sample variance $\sigma_{\mathrm{blk}}^2$ are:
\begin{equation}
\mu_{\mathrm{blk}}=\frac{1}{n_{\mathrm{blk}}}\sum_{j=1}^{n_{\mathrm{blk}}}c_j,
\qquad
\sigma_{\mathrm{blk}}^2=
\begin{cases}
\displaystyle
\frac{1}{n_{\mathrm{blk}}-1}\sum_{j=1}^{n_{\mathrm{blk}}}(c_j-\mu_{\mathrm{blk}})^2,
& n_{\mathrm{blk}}>1,\\[6pt]
0,&n_{\mathrm{blk}}=1.
\end{cases}
\end{equation}
The number of matching records $n_{\mathrm{blk}}$ is not an independent sample size: credits may come from the same trajectory.

For $n_{\mathrm{blk}}>0$, the scaling factor is $r_{\mathrm{blk}}(s,a)=n_{\mathrm{blk}}/(n_{\mathrm{blk}}+\nu)$; we set $r_{\mathrm{blk}}(s,a)=0$ when no records match. When $\nu=0$ and at least one record matches, $r_{\mathrm{blk}}(s,a)=1$. The centering mean $V_{\mathrm{blk}}(s)$ averages the block-wise credit estimates $Q_{\mathrm{blk}}(s,a)$ over candidate actions $\mathcal A(s)$:
\begin{equation}
V_{\mathrm{blk}}(s)=
\begin{cases}
\displaystyle
\frac{\sum_{a\in\mathcal A(s)}r_{\mathrm{blk}}(s,a)Q_{\mathrm{blk}}(s,a)}
{\sum_{a\in\mathcal A(s)}r_{\mathrm{blk}}(s,a)},
& \displaystyle\sum_{a\in\mathcal A(s)}r_{\mathrm{blk}}(s,a)>0,\\[10pt]
0,&\text{otherwise}.
\end{cases}
\end{equation}

For $n_{\mathrm{blk}}=0$, we set $Q_{\mathrm{blk}}(s,a)=0$ without evaluating Equation~\ref{eq:block_value}. For $n_{\mathrm{blk}}=1$, we set the sample variance to zero, so only the scaling factor $r_{\mathrm{blk}}$ reduces the estimate. A single credit does not establish that later executions will receive the same credit. The variance correction in Equation~\ref{eq:block_value} reduces the magnitude of the block-wise credit estimate toward zero without reversing its sign.  We fix $\nu=3$, so three matching records receive a count-based scaling factor of $0.5$. A single record can still contribute, but receives less weight, which increases as more records accumulate. We fix $\xi=0.5$ to account for variation among credit records. For the same mean, variance, and record count, this is more conservative than no variability correction ($\xi=0$), while retaining at least as much credit magnitude as a unit correction ($\xi=1$).

\subsection{Step-wise Credit Estimates}
\label{app:step_value}

Let $\mathcal N(s)$ denote the step-wise credit records retrieved for execution context $s$, including records from similar observations when semantic retrieval is enabled. Let $\mathcal N(s,a)$ denote the subset whose actions match candidate action $a$. For record $j$, let $C_j^{\mathrm{step}}$ be its assigned step-wise credit. We multiply each credit by a scaling factor $q_j$:
\begin{equation}
q_j=
\begin{cases}
q_{\mathrm{replay}},
& \text{episode designated for route replay},\\
1,
& \text{other episode, with progress beyond its initial frontier},\\
q_{\mathrm{pre}},
& \text{otherwise}.
\end{cases}
\end{equation}
Progress beyond the episode's initial frontier means that the total score immediately before or after the action exceeds the progress frontier before that episode's memory update. The configured factors $q_{\mathrm{replay}}$ and $q_{\mathrm{pre}}$ scale credits from replay episodes and other credits at or below the frontier, respectively. Credits from replay episodes receive $q_{\mathrm{replay}}$ regardless of total score. For nonempty $\mathcal N(s)$, the centering mean $V_{\mathrm{step}}(s)$ is
\begin{equation}
    V_{\mathrm{step}}(s)= \frac{\sum_{j\in\mathcal N(s)} q_jC_j^{\mathrm{step}}}{|\mathcal N(s)|}.
\end{equation}
The step-wise credit estimate $Q_{\mathrm{step}}(s,a)$ averages $q_jC_j^{\mathrm{step}}$ over $\mathcal N(s,a)$, whereas $V_{\mathrm{step}}(s)$ averages over all retrieved step-wise records $\mathcal N(s)$. Both denominators count records; neither denominator sums the scaling factors.

If no step-wise records match candidate action $a$, the step-wise gate $g_{\mathrm{step}}$ is zero and that action receives no step-wise contribution to the combined credit estimate $A_{\mathcal M}(s,a)$.

\paragraph{Support Gates.}
\label{app:credit_gates}
Support gates allow limited evidence to influence decisions while reducing its contribution when few records match. For ordinary exact-context block-wise retrieval, we use
\begin{equation}
g_{\mathrm{blk}}(s,a)=
\begin{cases}
0, & n_{\mathrm{blk}}=0,\\
\dfrac{n_{\mathrm{blk}}}{n_{\mathrm{blk}}+\kappa},
& 0<n_{\mathrm{blk}}<m_{\mathrm{blk}},\\
1, & n_{\mathrm{blk}}\ge m_{\mathrm{blk}},
\end{cases}
\end{equation}
with $m_{\mathrm{blk}}=2$ and $\kappa=0.8$. Thus, a single matching record receives a gate of approximately $0.56$, while two or more records receive a gate of $1$. This allows an initial observation to contribute without giving it the full channel weight.

For step-wise retrieval, let $n_{\mathrm{step}}=|\mathcal N(s,a)|$ be the number of matching records. We use
\begin{equation}
g_{\mathrm{step}}(s,a)
=\min\left(1,\frac{n_{\mathrm{step}}}{m_{\mathrm{step}}}\right)
\end{equation}
This gate increases linearly with the record count and reaches $1$ at three records. We set $m_{\mathrm{step}}=3$, giving gate values of $1/3$, $2/3$, and $1$ for one, two, and three matching records, respectively. This gradually increases the step-wise contribution as records accumulate. Both gates operate before standardization.

\paragraph{Algorithm, episode and block updates.}
\label{app:algo}
Algorithm~\ref{alg:earn} summarizes the update order.
\begin{algorithm}[t]
\caption{EARN}
\label{alg:earn}
\small
\begin{algorithmic}[1]
\Require Frozen LLM $f_\theta$; budgets $K,H$;
block size $B$; consolidation length $K_c$
\State Initialize $\mathcal M$, $\mathcal T\gets\emptyset$, $F\gets0$
\For{$k=1,\ldots,K$}
    \State Reset environment; initialize $s$ and $\tau_k\gets\emptyset$
    \State If $k\le K-K_c$ and $\mathcal T=\emptyset$, set $F_b\gets F$
    \State $\mathcal R^\star\gets\operatorname{RetrieveRoute}(\mathcal M)$;
    $m\gets1$
    \State $r\gets(k>K-K_c)\land(\mathcal R^\star\neq\emptyset)$

    \Statex \textcolor{blue!70!black}{\textit{Conditional replay or action-probability adjustment}}
    \For{up to $H$ steps, until termination}
        \State $r\gets r\land\operatorname{ReplayValid}(\mathcal R^\star,m,s)$
        \If{$r$}
            \State $a\gets a_m^\star$; $m\gets m+1$
        \Else
            \State $(\mathcal A,p^0)\gets
            \operatorname{ProposeAndNormalize}(f_\theta,s,\mathcal M)$
            \State Estimate $A_{\mathrm{blk}},A_{\mathrm{step}},
            g_{\mathrm{blk}},g_{\mathrm{step}},q_{\mathrm{step}}$
            from retrieved records
            \State $A_{\mathcal M}\gets
            g_{\mathrm{blk}}A_{\mathrm{blk}}
            +\lambda_{\mathrm{step}}[1-\eta g_{\mathrm{blk}}]_+
            g_{\mathrm{step}}q_{\mathrm{step}}A_{\mathrm{step}}$
            \State $\widehat A_{\mathcal M}\gets
            \operatorname{StandardizeAndClip}(A_{\mathcal M})$
            \State $p_{\mathcal M}(a\mid s)\propto
            p^0(a\mid s)\exp(\beta\widehat A_{\mathcal M}(s,a))$
            \State $a\gets\operatorname{Select}(p_{\mathcal M})$
        \EndIf
        \State Execute $a$; observe feedback and next context $s'$
        \State Append transition and execution metadata to $\tau_k$;
        $s\gets s'$
    \EndFor

    \Statex \textcolor{blue!70!black}{\textit{Every episode: update step-wise credit and routes}}
    \State $\mathcal C_k^{\mathrm{step}}\gets
    \operatorname{StepCredit}(\tau_k,F)$
    \State Update step-wise records in $\mathcal M$
    using $\mathcal C_k^{\mathrm{step}}$
    \State Update stored routes in $\mathcal M$ using $\tau_k$

    \Statex \textcolor{blue!70!black}{\textit{At block completion: select and attribute credit}}
    \If{$k\le K-K_c$}
        \State $\mathcal T\gets\mathcal T\cup\{\tau_k\}$
        \If{$|\mathcal T|=B$ or $k=K-K_c$}
            \State $\tau^{\mathrm{sel}}\gets\operatorname{SelectBest}(\mathcal T)$
            \State Update block-wise records in $\mathcal M$ with
            $\operatorname{BlockCredit}(\tau^{\mathrm{sel}},F_b)$
            \State $\mathcal T\gets\emptyset$
        \EndIf
    \EndIf
    \State Update $F$ using repeated recent positive route scores
\EndFor
\State \Return $\mathcal M$
\end{algorithmic}
\end{algorithm}

\subsection{Action Selection and Route Replay}
\label{app:action_retrieval}

\paragraph{Context and Action Matching.} Execution contexts combine normalized observations with supplementary state information and an environment-provided state identifier when available. Actions are matched after normalizing case, whitespace, and surrounding punctuation.

\paragraph{Retrieval from Similar Observations.} In Jericho and ScienceWorld, semantic retrieval supplements insufficient exact matches with records from similar observations. For block-wise retrieval, estimates from similar contexts are combined using observation-similarity weights and a transfer coefficient. Step-wise retrieval instead pools additional records before estimating action values. For Crafter, we disable this observation-text retrieval mechanism in our multimodal configuration and use exact-context retrieval for action evidence. Similarity-based retrieval does not replace the context checks required for route replay.

\paragraph{Memory-Based Probability Adjustment.}
\label{app:probability_adjustment}
We normalize the LLM-reported confidence scores. When memory candidates are added, we assign them initial weights and renormalize to obtain $p^0(a\mid s)$. Fused advantages are standardized using the candidate mean and population standard deviation, then clipped to $[-\zeta,\zeta]$. If the standard deviation is below $10^{-8}$, the divisor is replaced by the larger of one and the maximum absolute centered advantage. Base probabilities are floored at $10^{-12}$ before taking logarithms.  We fix $\zeta=3$ and clip standardized estimates to $[-3,3]$ before probability adjustment, bounding the credit-dependent exponent to $[-3\beta,3\beta]$. For a final candidate set of at most five actions, population-standardized estimates lie within $[-2,2]$, so this bound leaves them unchanged.

\paragraph{Route Storage and Replay.} Route memory is updated after each episode, independently of block-wise credit attribution. Candidate routes are compared by score, with shorter routes preferred when scores tie. Before each replayed action, the agent checks context agreement, action availability when a valid-action list is supplied, and stored failure-based restrictions. A failed check or exhausted route ends replay for the current episode, and the agent resumes memory-guided candidate generation and probability adjustment.

\section{Additional Experimental Settings}
\label{app:final_window}

\paragraph{Evaluation window and replay schedule.} We evaluate the final $W=4$ episodes and use blocks of $B=5$ episodes for block-wise credit attribution. The evaluation window and block size are separate protocol choices. Conditional route replay is enabled in the final four episodes. Step-wise credit attribution uses every episode, including episodes designated for replay.

\paragraph{Hyperparameters and Hardware.}
\begin{itemize}
\item \textbf{Discount factor ($\gamma=0.95$).} This setting allows scoring-event feedback to reach earlier preparatory actions while decaying with temporal distance. An action occurring ten steps before a scoring event receives that event's signal with a discount weight of $0.95^{10}\approx0.60$.

\item \textbf{Probability-adjustment strength ($\beta=0.75$).} This coefficient controls how strongly memory changes the base action probabilities. When two candidates differ by one unit in their standardized and clipped credit estimates, the adjustment multiplies their base probability ratio by $e^{0.75}\approx2.12$ in favor of the higher-valued candidate. Memory can therefore meaningfully influence selection while the LLM's base distribution continues to contribute to the adjusted probabilities.
\item \textbf{Block-wise credit parameters ($\alpha_{\mathrm{pre}}=0.02$, $\rho=0.5$).} For repeated-progress segments, credits follow $0.02i/L_e$, reaching 0.02 at the scoring action. For new-progress segments, credits follow $0.5+0.5i/L_e$, reaching 1 at the scoring action. This assigns higher credit to new progress while retaining small positive credits for repeated progress.

\item \textbf{Base step-wise weight ($\lambda_{\mathrm{step}}=0.35$).} The base weight of step-wise credit is fixed at 0.35. Its contribution to the combined estimate is further regulated by block-wise support, the step-wise support gate, and record scaling.

\item \textbf{Step-wise suppression coefficient ($\eta=0.85$).} The block-wise support gate $g_{\mathrm{blk}}(s,a)\in[0,1]$ modulates the step-wise coefficient:
\[
\lambda_{\mathrm{step}}
[1-\eta g_{\mathrm{blk}}(s,a)]_+
=
0.35[1-0.85g_{\mathrm{blk}}(s,a)]_+.
\]
This coefficient decreases from 0.35 when $g_{\mathrm{blk}}=0$ to 0.0525 when $g_{\mathrm{blk}}=1$, before applying the step-wise gate $g_{\mathrm{step}}$ and record scale $q_{\mathrm{step}}$. Thus, stronger block-wise support attenuates the step-wise contribution before standardization without eliminating it through suppression alone.
\end{itemize}
For EARN, the generation temperature is 0.8, and the LLM is requested to propose five candidate actions per decision. We serve the W4A16-quantized Gemma-4-31B-it checkpoint using vLLM on a single NVIDIA H100 GPU. Model parameters remain frozen throughout evaluation.

\section{Inference Savings from Conditional Replay}
\label{app:replay-savings}

\paragraph{Measurement.} We measure how conditional replay reduces LLM candidate-generation calls on the Jericho runs reported in Table~\ref{tab:jericho-last4}. A replayed action is still executed in the environment, but bypasses LLM candidate generation. Thus, each replayed step avoids one logical generation call without reducing the number of environment interactions. We count replayed steps from transition logs and verify that non-replay steps match the corresponding decision-history records. This measurement excludes API retries, token consumption, and latency.

\begin{table}[!htbp]
\centering
\small
\caption{Candidate-generation call savings (\%) for \method on five Jericho games. Last-4 averages the final four per-episode rates; Last reports the final episode. Final/Run counts only replayed steps in the final four episodes, divided by all executed steps in the run. Average gives equal weight to each game. Budget labels give episodes/maximum steps per episode.}
\vspace{1mm}
\label{tab:jericho-inference-savings-detailed}
\setlength{\tabcolsep}{5pt}
\renewcommand{\arraystretch}{1.1}
\begin{tabular}{@{}l*{9}{r}@{}}
\toprule
\textbf{Game}
& \multicolumn{3}{c}{\textbf{30/30}}
& \multicolumn{3}{c}{\textbf{50/50}}
& \multicolumn{3}{c}{\textbf{70/70}} \\
\cmidrule(lr){2-4}
\cmidrule(lr){5-7}
\cmidrule(lr){8-10}
& \textbf{Last-4} & \textbf{Last} & \textbf{Final/Run}
& \textbf{Last-4} & \textbf{Last} & \textbf{Final/Run}
& \textbf{Last-4} & \textbf{Last} & \textbf{Final/Run} \\
\midrule
DeepHome
& 93.08 & 100.00 & 15.84
& 84.88 & 84.00 & 9.05
& 90.00 & 90.00 & 5.22 \\
Detective
& 100.00 & 100.00 & 14.13
& 97.87 & 97.87 & 9.24
& 97.22 & 97.22 & 5.46 \\
Zork I
& 76.67 & 90.00 & 10.98
& 88.00 & 88.00 & 8.22
& 78.28 & 90.91 & 5.35 \\
Zork III
& 84.85 & 90.91 & 10.67
& 50.55 & 48.65 & 4.21
& 92.69 & 82.76 & 3.96 \\
ZTUU
& 73.33 & 73.33 & 10.32
& 73.23 & 82.00 & 8.07
& 73.17 & 50.00 & 4.72 \\
\midrule
\textbf{Average}
& 85.59 & 90.85 & 12.39
& 78.90 & 80.10 & 7.76
& 86.27 & 82.18 & 4.94 \\
\bottomrule
\end{tabular}
\end{table}

\paragraph{Metrics.} For game $g$, let $T_{g,k}$ denote the number of executed steps in episode $k$, and let $N^{\mathrm{replay}}_{g,k}$ denote the number of directly replayed steps. Table~\ref{tab:jericho-inference-savings-detailed} reports three percentages:
\begin{equation}
\begin{aligned}
\mathrm{Save}_g^{\mathrm{Last\text{-}4}} &= \frac{100}{4}\sum_{k=K-3}^{K}\frac{N^{\mathrm{replay}}_{g,k}}{T_{g,k}},\\
\mathrm{Save}_g^{\mathrm{Last}} &= 100\frac{N^{\mathrm{replay}}_{g,K}}{T_{g,K}},\\
\mathrm{Save}_g^{\mathrm{Final/Run}} &= 100\frac{\sum_{k=K-3}^{K}N^{\mathrm{replay}}_{g,k}}{\sum_{k=1}^{K}T_{g,k}}.
\end{aligned}
\end{equation}
Last-4 averages the per-episode saving rates over the final four episodes, and Last reports the final episode's rate. Final/Run divides replayed steps in the final four episodes by all executed steps in the run. Each reported average gives equal weight to the five games.

\paragraph{Results.} Conditional replay avoids 78.90--86.27\% of candidate-generation calls in the final four episodes, averaged equally across games. These final-window savings correspond to 12.39\%, 7.76\%, and 4.94\% of all executed steps under the 30/30, 50/50, and 70/70 budgets, respectively. When replay becomes inapplicable, the agent resumes LLM candidate generation and probability adjustment. These counts measure the generation calls bypassed during the recorded runs. They do not measure how episode scores would change if the agent generated an action at every step.

\section{Case Study}

\begin{figure}[!htbp]
\centering
\begin{tikzpicture}[
  font=\small,
  panel/.style={
    draw=black!35, rounded corners=2pt, inner sep=7pt,
    align=left, text width=.41\textwidth
  },
  wide/.style={
    draw=black!35, rounded corners=2pt, inner sep=7pt,
    align=left, text width=.89\textwidth
  },
  flow/.style={-{Latex[length=2mm]}, thick, draw=black!55}
]
\node[panel, fill=blue!6, draw=blue!45] (block) {%
\textbf{Block-wise credit: Episodes 6--10}\par
Episode 6 is selected. Its progress-bearing segment contains \texttt{north} $\rightarrow \cdots \rightarrow$ \texttt{take mask}. Credit extends to preparatory actions in that segment.
};

\node[panel, fill=orange!7, draw=orange!55,
      right=7mm of block] (step) {%
\textbf{Step-wise credit: Episode 7}\par
Although Episode 7 is not selected, its feedback is retained. In the same context, \texttt{north} has discounted return $G=2.858$.
};

\coordinate (sources) at ($(block.south)!0.5!(step.south)$);
\node[wide, fill=green!6, draw=green!45,
      below=12mm of sources] (decision) {%
\textbf{Later decision: Episode 12, step 2}\par
In the matching context, combined memory reinforces \texttt{north}: its candidate probability changes from $\mathbf{0.600 \rightarrow 0.763}$, and the agent selects it.
};
\draw[flow] (block.south) -- ($(decision.north)+(-2.2cm,0)$);
\draw[flow] (step.south) -- ($(decision.north)+(2.2cm,0)$);

\node[wide, fill=purple!6, draw=purple!45,
      below=10mm of decision] (final) {%
\textbf{Final-window replay: Episode 30}\par
The stored route from \textbf{Episode 25} is replayed at steps 0--21. At step 22, an execution check fails and candidate generation resumes: \texttt{take bow} $\rightarrow$ \texttt{take red lens} $\rightarrow \cdots \rightarrow$ \texttt{take coin}. \textbf{Final score: 22.}
};
\end{tikzpicture}
\caption{ZTUU case study in the 30-episode, 30-step setting. The upper panels illustrate the two credit sources and a later combined-memory decision. The lower panel separately shows conditional replay and fallback in the final episode. The probability change reflects accumulated memory, not an isolated contribution from either displayed episode.}
\label{fig:ztuu-case}
\end{figure}
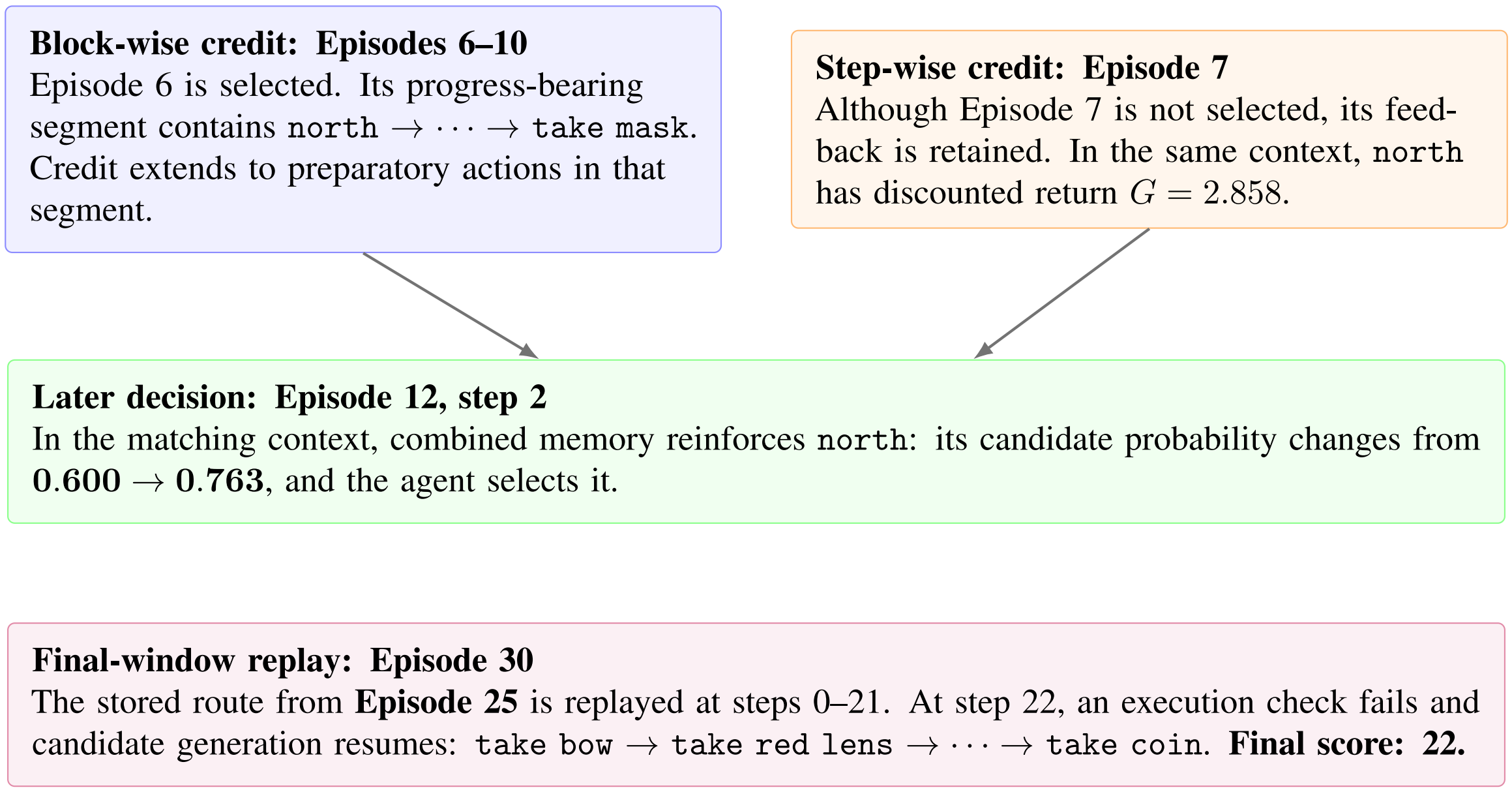

\paragraph{Case analysis.} Figure~\ref{fig:ztuu-case} shows that the key to conditional replay is that it reuses a validated route prefix without locking the agent into the entire stored trajectory. Block-wise credit emphasizes progress in high-performing trajectories, while step-wise credit retains feedback from other episodes; when replay no longer applies, the agent can return to memory-guided action generation. ZTUU's final episode illustrates this flexibility. The first 22 replayed steps bring the score to 12. After the replay diverges at step 22, the agent resumes generation, obtains further progress through \texttt{take red lens} and \texttt{take coin}, and recovers its previous best score of 22.

Thus, replay preserves a reliable prefix while leaving the remaining steps open to adaptation, rather than forcing the agent to repeat an entire old route. The avoided LLM calls could potentially be reallocated to additional exploration under a fixed call budget, provided further environment interactions are available; this reallocation is not evaluated here.

\section{Additional Controlled Comparisons}

\subsection{Baselines with Their Native Implementations}
We run Reflexion, AWM, EvoTest, and JitRL following the protocols in their publicly available implementations. Each retains its method-specific decoding, memory, and retrieval settings. For FIFO Memory, we prepend full transcripts of previous episodes to the prompt and remove the oldest transcripts when the memory context budget is exceeded. Within each benchmark, all methods use the same frozen backbone, task instances, interaction budgets, and scoring protocol. This preserves their distinct decision mechanisms while controlling the model and evaluation conditions.

\subsection{EARN under the Matched Evaluation Protocol}
EARN uses the same backbone and evaluation protocol as the baselines, without parameter updates. For EARN, we set the action-generation temperature to 0.8 across all three benchmarks, five candidate actions, and all-MiniLM-L6-v2 embeddings for semantic retrieval.

\subsection{Prompt Template}
EARN programmatically retrieves historical action--outcome records from persistent memory and formats them into the prompt, without additional LLM summarization calls. The following simplified template illustrates the structure used for candidate-action generation.

\begin{figure}[!htbp]
\centering
\begin{tcolorbox}[
    width=\linewidth,
    colback=white,
    colframe=blue!45!black,
    colbacktitle=blue!8,
    coltitle=black,
    title={EARN Action Generation Prompt},
    fonttitle=\bfseries,
    boxrule=0.6pt,
    arc=1mm,
    left=8pt, right=8pt,
    top=6pt, bottom=6pt
]
\small
\textbf{\textcolor{blue!60!black}{System:}}\par
You are an agent interacting with an environment. Use the current observation and past experience to propose actions that make progress toward the task objective.

\medskip
\textbf{Task:} \texttt{\{task\_description\}}

\medskip
\textbf{Current observation:}\par
\texttt{\{observation\}}

\medskip
\textbf{Recent interaction history:}\par
\texttt{\{recent\_history\}}

\medskip
{\setlength{\fboxsep}{6pt}%
\colorbox{green!7}{%
\parbox{\dimexpr\linewidth-2\fboxsep\relax}{%
\textbf{\textcolor{green!40!black}{Cross-episode memory:}}\par
The agent's memory module retrieved the following records for the current or similar execution contexts. They describe previously attempted actions and their observed outcomes. Consider their relevance to the current situation when proposing actions.\par
\smallskip
\texttt{\{retrieved action--outcome records\}}
}}}

\medskip
\textbf{Available actions, when provided:}\par
\texttt{\{available\_actions\}}

\medskip
\textbf{\textcolor{blue!60!black}{Output instruction:}}\par
Propose five distinct candidate actions with brief reasoning. Assign each candidate action an integer confidence score from 0 to 100 indicating your relative preference for selecting that action. The five scores must sum to 100. Follow the provided action constraints.

\end{tcolorbox}
\caption{EARN general prompt structure. Observations are textual in Jericho and ScienceWorld and include visual input in Crafter.}
\label{fig:earn_prompt}
\end{figure}

\end{document}